\documentclass[11pt]{article}
\usepackage[paper=letterpaper,centering,margin=0.8in]{geometry}

\usepackage{amssymb,amsmath,amsthm}
\usepackage{authblk}

\newtheorem{theorem}{Theorem}

\newtheorem{lemma}{Lemma}
\newtheorem{proposition}{Proposition}

\usepackage{bm}
\usepackage{xcolor}
\usepackage[ruled,vlined]{algorithm2e}

\usepackage{dsfont}
\usepackage{bbm}
\usepackage{bm}
\usepackage[mathscr]{eucal}
\usepackage{mathtools}
\usepackage[]{color}
\usepackage{tabularx}
\newcolumntype{Y}{>{\centering\arraybackslash}X}

\usepackage{caption}
\usepackage{subcaption}
\usepackage{mwe}
\usepackage{tikz}
\usetikzlibrary{positioning}
\usepackage[ruled]{algorithm2e}
\usepackage{natbib}
 \bibpunct[, ]{(}{)}{,}{a}{}{,}%

\newenvironment{Repeatlemma}[1]
  {\renewcommand{\thelemma}{\ref{#1}}%
   \addtocounter{lemma}{-1}%
   \begin{lemma}}
  {\end{lemma}}
 \newenvironment{Repeatproposition}[1]
  {\renewcommand{\theproposition}{\ref{#1}}%
   \addtocounter{proposition}{-1}%
   \begin{proposition}}
  {\end{proposition}}

\usepackage{enumerate}

\def\v{{\bm v}}

\def\={\!=\!}

\usepackage[normalem]{ulem}

\newcommand{\Beginproof}{\begin{proof}}
\newcommand{\Endproof}{\end{proof}}

\begin{document}

\title{An Adiabatic Theorem for Policy Tracking with TD-learning}

\author{Neil Walton}
\affil{University of Manchester\\
\small{\texttt{neil.walton@manchester.ac.uk}}}

\maketitle

\abstract{
We evaluate the ability of temporal difference learning to track the reward function of a policy as it changes over time.
Our results apply a new adiabatic theorem that bounds the mixing time of time-inhomogeneous Markov chains.
%
%
We derive finite-time bounds for tabular temporal difference learning and $Q$-learning when the policy  used for training changes in time. 
To achieve this, we develop bounds for stochastic approximation under asynchronous adiabatic updates.
}

\maketitle

\section{Introduction}
Policy evaluation and, in particular, temporal difference (TD) learning is a key ingredient in reinforcement learning.
Here the expected value of future rewards is estimated from simulations of a given policy.
When a stationary policy is fixed, the simulated process is a time-homogeneous Markov chain.
The convergence of the policy evaluation algorithm is analyzed using stochastic approximation techniques under asynchronous Markovian updates.

There is a well-developed theory of stochastic approximation that establishes the convergence of a variety of policy evaluation schemes.
However, less attention has been placed on the convergence of the Markov chain that is estimated.
Usually theoretical results analyze a time-homogeneous Markov chain, which corresponds to evaluating a fixed policy. A mixing time condition is usually an assumption made in advance which specifies the rate of convergence to the chain's equilibrium distribution. 
For irreducible time-homogenous Markov chains geometric mixing often holds. 
However, when training a reinforcement learning algorithm the policy of interest is rarely fixed. 
Usually, the policy evaluation algorithm is combined with a policy improvement step.
Here the set of actions taken by the policy are updated and consequently the transition matrix and equilibrium distribution of the policy of interest evolves in time. 
This essential component of the reinforcement learning problem, in turn, invalidates the time-homogeneous assumption and mixing time conditions assumed in the theory of policy evaluation. 
In this work, we do not investigate the policy improvement mechanisms used in reinforcement learning.
Instead, we wish to understand the ability of policy evaluation algorithms to accurately assess a current policy as that policy is changed in time according to some separate mechanism. 

One should expect --as it is implicitly assumed-- that a good policy evaluation algorithm such as TD learning should be able to track the current policy as it changes in time. Thus, finite-time bounds derived in recent years for policy evaluation, should be applicable for policy tracking.
In this article, we analyze the policy tracking problem.  That is we fix a sequence of transition matrices and we analyze how effectively TD learning tracks a policy as it evolves and converges on some final policy.
%

A key component of our analysis is a new adiabatic theorem for time-inhomogeneous Markov chains. 
Adiabatic results are commonly used in physics (see \cite{griffiths2017introduction}).
These results study the Hamiltonian subject to changing external conditions, and prove that if the evolution is sufficiently slow, then the system state will be close to the ground state of the final Hamiltonian. Dynamics with this property are known as adiabatic. However when subjected to rapidly varying (diabatic) conditions there is insufficient time for the functional form to adapt and thus the state of the system is influenced by its initial configuration.
%
%
Adiabatic results have had a key role to play in the learning and stability of stochastic systems, see \cite{rajagopalan2009network}, and
results of this type are clearly applicable in the context of stochastic approximation and reinforcement learning. 
If we slowly update our policy over time then our time-inhomogeneous Markov chain should remain close to equilibrium and a temporal difference algorithm should be able to successfully evaluate the current policy as it evolves. However, if the policy varies too quickly then these changes will begin to dominate the rate of convergence of the policy evaluation algorithm.

Although the intuition above is clear, new results are required to quantify this convergence. 
Here we prove a general adiabatic theorem for irreducible, finite state-space, time-inhomogeneous Markov chains. We prove that so long as the component-wise change in transition probabilities goes to zero then the time-inhomogeneous chain remains close to the stationary distribution the most recent transition matrix.
This theoretical result on the convergence of time-inhomogeneous Markov chains may well be of independent interest. 

 As an application, we
show how our results can be applied to 
asynchronous stochastic approximation and thus TD-learning and Q-learning. The problem setting and our findings can be informally described as follows.
We consider an asynchronous Robbins-Monro scheme, $R_t$, with learning rate $\alpha_t = t^{-\gamma_{\alpha}}$ with $\gamma_{\alpha}>0$. We let $P^{(t)}$ be a sequence of Markov transition matrices for a time-inhomogeneous Markov chain. Similar to $\gamma_\alpha$, we let $\gamma_P$ of the rate of change in the entries of these matrices, and we let $\gamma_{\pi}$ be the rate of change in the smallest probability state (under the equilibrium distribution of $P^{(t)}$). 
Under the asynchronous scheme, the components of, $R_t$, are updated according to the transitions of this time-inhomogeneous chain.
We assess the ability of this scheme to approximate a fixed point $R^{\star}_T$ which depends on $P^{(T)}$. We find a bound of the form
\begin{align*}
  || R_T - R^{\star}_T||_\infty = O\left(  \frac{1}{T^{\gamma_P-\gamma_\alpha - \gamma_{\pi}}} + \frac{\sqrt{\log T}}{T^{\gamma_\alpha/2- 3 \gamma_\pi/2}}\right)\, .
\end{align*}
That is under the conditions $\gamma_P> \gamma_\alpha + \gamma_{\pi}$ and $\gamma_\alpha > 3 \gamma_\pi$, the chain is adiabatic and the stochastic approximation scheme provides a good assessment of the performance of the current policy.
These result transfer over in a straight-forward manner to give convergence guarantees for tabular $TD$-learning and $Q$-learning, under time-inhomogeneous change.
Although our goal is not to assess a static policy, we note that if the Markov chain's transition matrix $P$ does not change in time, then $\gamma_P=\infty$ and $\gamma_{\pi}=0$ and we obtain a bound of order $O(\sqrt{\log T}/T^{\gamma_{\alpha}/2})$ which is consistent with the best performance bounds for off-policy TD-learning and $Q$-learning (see \cite{qu2020finite}). From this we, further, see that if $\gamma_P-\gamma_\alpha-\gamma_\pi > (\gamma_\alpha-3\gamma_\pi)/2$ then the order of the convergence rate is the same as for a time-homogeneous learning problem.

The results above are considered for stochastic approximation and tabular reinforcement learning. However, in future work, we will show the framework above can be developed using different methodology applicable to online convex optimization and consequently applicable to linear function approximation frameworks in reinforcement learning. 

\subsection{Relevant Literature.}
Convergence of the algorithm for tabular learning was first established by \cite{sutton1988learning}. A general proof for $TD(\lambda)$ is given by  \cite{dayan1992convergence}. 
Convergence results generally require the convergence of an asynchronous stochastic approximation scheme. \cite{Tsitsiklis1994} provides a general criteria for the convergence of a fully asynchronous stochastic approximation scheme. However, more commonly, updates are considered to be formed from a time-homogeneous Markov chain. Here the book of 
\cite{benveniste2012adaptive} provides a general treatment. 

More recently there has been an increased focused on finite-time bounds for these schemes as a means of comparing and quantifying performance. The work of \cite{Bach2011} provides a relatively general convergence result for stochastic approximation in the setting of convex optimization. Such results can then be applied in the context of stochastic approximation and reinforcement learning. The paper of \cite{qu2020finite} provides a recent instance of this approach and we develop parts of their analysis to derive our bounds for tabular temporal difference learning. A distinct but not entirely unrelated area of interest off-policy evaluation. This considers a different setting where there is also a disparity between the Markov chain used for training and target that is to be evaluated. See \cite{precup2000eligibility} and \cite{duan2020minimax} for two key references in the tabular and function approximation setting.

   
The works discussed above tend to focus on finite-time bounds under independent or time-homogeneous Markov feedback. 
Here results on mixing times are generally required. Mixing times results for time-homogeneous Markov chains are given in the following texts \cite{aldous-fill-2014,levin2017markov,montenegro2006mathematical}.
However as discussed, in reinforcement learning, we often cannot assume the time homogeneity of our Markov chain.

Thus we wish to focus on time-inhomogeneous Markov input. Here we require a result that will replace common-place mixing time assumptions. 
Adiabatic results focus on the ability of a time-inhomogeneous Markov chain to approximate the stationary distribution of a time-homogeneous chain. The results of \cite{bradford2011adiabatic} and  \cite{rajagopalan2009network} give two such examples for a time-inhomogeneous Markov chains in specific applications areas. 
 Our analysis relies on the results of \cite{seneta1988perturbation,seneta1993sensitivity,seneta2006non} who derives contraction properties in order to develop a sensitivity analysis of time-inhomogeneous Markov chains. Also see the review of \cite{ipsen2011ergodicity}. We apply this style of analysis along with recursions similar to those in stochastic approximation to establish our bounds. These are then applied to temporal difference learning to develop the finite-time bounds. 

\subsection{Outline.} In Section \ref{sec:model}, we present our main mathematical models, assumptions and notation.
 In Section \ref{sec:adabiatic}, we prove our Adiabatic Theorem. Here Theorem \ref{thrm:adiabatic} gives a mixing time result for time-inhomogeneous Markov chains. 
In Section \ref{sec:async}, we first apply this to asynchronous stochastic approximation in Theorem \ref{thrm:asych}.
In Section \ref{sec:TD}, we the use this to study the problem of policy tracking for tabular TD(0) learning.
%
%

\section{Model and Assumptions.}\label{sec:model}

Below in Section \ref{sec:Notation}, we give several definitions for vectors, probability distributions and distances between them.
Then, in Section \ref{sec:MC}, we give definitions for time inhomogeneous Markov chains including the definition of the coefficient of ergodicity, which is an object of central interest in the paper. In Section \ref{TabTD}, we define temporal difference learning and in Section \ref{sec:TabQ} we define Q-learning.

\subsection{Mathematical Definitions and Notation.}
\label{sec:Notation}
We let $\mathbb Z_+=\{0,1,2,...\}$ denote the non-negative integers.
We define $\bm 1 = (1 : x \in \mathcal X)$ to be the vector of all ones and $\bm e_x$ to be the $x$-th unit vector. For two probability distributions, $\lambda$ and $\pi$, defined on finite state space $\mathcal X$, the total variation distance between $\lambda$ and $\pi$ is defined by \footnote{Often total variation distance has the subscript $TV$, i.e. $|| \mu -\pi ||_{TV}$. However, we will be primarily be using this norm and so omit the subscript.}
\[
||\lambda - \pi ||
:=
\frac{1}{2}\sum_{x\in \mathcal X} | \lambda_x - \pi_x | \, .
\] 
For two matrices $P$ and $P'$, we define
\[
|| P - P'|| = \max_{x\in\mathcal X} || P_{x\cdot} - P'_{x\cdot} || \,.
\]
Specifically for the two Markov chain transition matrices (as   below in Section \ref{sec:MC}), this is the maximum total variation distance between the transitions. 

We let $|| \cdot ||_\infty$ denote the supremum norm, that is $|| z ||_\infty = \max_{x\in\mathcal X} |z_x| $ for $z \in \mathbb R^{\mathcal X}$. We say that an operation $z \mapsto F z$ where $Fz \in\mathbb R^{\mathcal X}$ is a $\beta$-contraction with respect to the supremum norm if 
\[
|| Fx - Fx' ||_{\infty} \leq \beta || x- x' ||_\infty\, .
\]
%
We will frequently use the subscript $\max$ (respectively $\min$) to denote the modulus of the largest  (resp. smallest) element in a set. For instance if $\theta \in \Theta$ denotes a set of parameters then 
$
\theta_{\max} = \max_{\theta \in \Theta} ||\theta||
$ and, for probability distribution $\pi$, $\pi_{\min}$ is the smallest probability event, $\pi_{\min}= \min_{x\in\mathcal X} \pi_x$.
%

\subsection{Markov chains}\label{sec:MC}
We discuss Markov chains and their mixing times properties. \cite{montenegro2006mathematical} provides a good general treatment of the mixing time of non-reversible Markov chains. 

We let $(\hat x_t : t \in \mathbb Z_+)$ be a discrete time Markov chain with states in the finite set $\mathcal X$. We let $n=|\mathcal X|$. 
The transition matrix of a (time-homogeneous) Markov chain is a non-negative matrix, $P$, whose rows sum to one, i.e. $\sum_{y\in\mathcal X} P_{xy}=1$ for all $ x\in \mathcal X$. 
We consider time-inhomogeneous Markov chains, that is a Markov chain whose transition matrix changes in time. 
We let $P^{(t)}$ denote the transition matrix of the $t$-th transition of the Markov chain $\hat x$. Thus the Markov chain $\mathcal X$ obeys the Markov property:
\[
\mathbb P (
\hat x_{t+1} = y | \hat x_t = x, \hat x_{t-1},...,\hat x_0)
= 
\mathbb P(\hat x_{t+1} = y | \hat x_t =x )
=
P^{(t)}_{xy}\, .
\]
Throughout the paper we will apply the convention that $\mathbb E_x [f(\hat x) ] = \mathbb E [f(\hat x_1) | \hat x_0 = x]$.
We assume each transition matrix, $P$ (and $P^{(t)}$), is irreducible meaning that there is a positive probability of transitioning between any pair of states over some finite-time. 
It is known that for finite-time chains that irreducibility implies there is a unique probability distribution $ \pi= (\pi_x: x \in \mathcal X)$ satisfying 
\[
 \pi P = \pi \, .
\]
This is called the equilibrium distribution (or stationary distribution) of $P$.
All eigenvalues of the matrix must, necessarily, be less than or equal to $1$. Thus irreducibility implies that the modulus of the second largest eigenvalue, $\rho_2$, is less than $1$.

Since the distribution at time $t$ of a time-homogeneous Markov chain evolves according to power of the matrix $P$, the 2nd largest eigenvalue determines the rate of convergence to equilibrium, specifically results of the form 
\[
||\mathcal P ( \hat x_T \in \cdot ) -  \pi (\cdot) || \leq C \rho_2^T \, , 
\]
are common-place. For instance, see Proposition 2.12 of \cite{montenegro2006mathematical}. 
The bounds such as the above holds for time-homogeneous (reversible) Markov chains, whereas we wish to consider time-inhomogeneous chains, which are not reversible. In this case the spectrum of the empirical transition matrix is not tractable. We instead consider the \emph{coefficient of ergodicity}, which is defined as follows,
\begin{equation}\label{eqn:ergodic}
	\rho(P)
:=
\sup_{
\substack{ \lambda: ||  \lambda|| =1\\  \lambda \cdot  \bm 1 = 0}  }
 || \lambda P ||
\end{equation}
Shortly, in Proposition \ref{prop:ergprop}, we will see that 
$\rho(P)$ is the total variation between the rows of $P$:
\[
\rho(P) :=\max_{x,x'} \big\| P_{x} - P_{x'} \big\|\,.
\]
There are several key properties of the coefficient of ergodicity that make it a good alternative to the modulus of the 2nd largest eigenvalue. These known results are summarized shortly in Proposition \ref{eqn:ergodic} in the next section.

\subsection{Tabular Temporal Difference Learning}\label{TabTD}
For a Markov chain $(\hat x_t: t\in\mathbb Z_+)$ with transition matrix $P$, the aim of temporal difference algorithm is to estimate the reward function
\begin{equation} \label{reward_function}
R(x) = R(x;P) := \mathbb E_x \left[
\sum_{t=0}^\infty \beta^t r(\hat x_t)
\right]	
\end{equation}
where $r:\mathcal X \rightarrow \mathbb R$ is a bounded instantaneous reward function and $\beta \in (0,1)$. 
The transition probabilities of the Markov chain are assumed to be unknown, we seek to calculate $R(x)$ by sampling from the Markov chain. Monte-carlo simulation will work but requires an entire sample path and can have high variance. So an alternative is to bootstrap from past estimates. Specifically, $R(x)$ satisfies the identity
\begin{equation}
	0= r(x) + \beta \mathbb E_x [R(\hat x_1)] - R(x).
\end{equation}
Thus we can seek a fixed point through an asynchronous Robbins-Monro scheme:
\begin{align*}
R_{t+1}(\hat x_t) =
R_{t}(\hat x_t) 	
+
 \alpha_t 
\left[
r(\hat x_t) 
+ \beta 
R_t(\hat x_{t+1}) - R_t(\hat x_t)
\right]
\end{align*}
and $R_{t+1}(x) = R_t(x)$ for all $x\neq \hat x_t$. A proof of convergence is given by \cite{Tsitsiklis1994}. The above algorithm is known as $TD(0)$. There are other variants of this algorithm such as $TD(\lambda)$ and $n$-step $TD$ algorithms. In this paper we focus on $TD(0)$. The analysis given in the paper will, almost certainly, transfer to these cases however for concreteness we focus on $TD(0)$. 
A key property is that the operator $F$ defined by 
\begin{equation}\label{Fcontract}
	FR(x) = r(x) + \beta \mathbb E_x [ R(\hat x)]  \,
\end{equation}
is a $\beta$-contraction with respect to the supremum norm: that is for all $R= (R(x) : x \in \mathcal X)$ and $R' = ( R'(x) : x\in\mathcal X)$ it holds that
\[
|| FR - FR' ||_\infty \leq \beta || R - R' ||_\infty \, .
\]
(see Lemma \ref{lem:contraction} in the appendix for a proof.)
We let $F_t$ be the transition operator defined by \eqref{Fcontract} where the expectation is taken with respect to the transition matrix $P^{(t)}$.

\subsection{Tabular Q-learning}\label{sec:TabQ}
We now consider a Markov chain that chooses states $s \in \mathcal S$ and actions $a \in \mathcal A$ according to a Markov chain $\hat x = ((\hat s_t, \hat a_t) : t\in\mathbb Z_+)$. We let $P^{(t)}$ be the transition matrix of this chain. That is $P^{(t)}_{(s,a),(s',a')}$ is the probability of taking action $a'$ in state $s'$ after being in state $s$ and taking action $a$ at time $t$.

Given a transition matrix $P$, a bounded function $r(s,a)$ and constant $\beta\in (0,1)$, the task of $Q$-learning is to evaluate the optimal $Q$-factor, which satisfies the fixed point equation
\[
0= r(s,a) + \beta \mathbb E_{s,a} \Big[ \max_{ a'\in\mathcal A} Q(  \hat s, a')\Big] -Q(s,a) 
\]
where here $\hat s$ is first state reached under transition Matrix $P$ after taking action $a$ in state $s$. Like with the reward function $R(x;P)$ in \eqref{reward_function}, we sometimes wish to make explicit the dependence on the transition matrix, $P$, in which case we write $Q(s,a;P)$

Like TD-learning, $Q$-learning is an asynchronous Robbins-Monro scheme defined as follows:
\[
Q_{t+1} (\hat s_{t}, \hat a_{t}) 
= 
Q_t(\hat s_{t}, \hat a_{t}) 
+
 \alpha_t 
\left[
r(\hat s_t,\hat a_t) 
+ \beta 
\max_{a\in\mathcal A} Q_t(\hat s_{t+1},a) - Q_t(\hat s_t, \hat a_t)
\right] \, .
\]
and $Q_{t+1}(s,a) = Q_t(s,a)$ for all $(s,a)$ such that $s\neq \hat s_t$ or $a \neq \hat a_t$. Again a key property of $Q$-learning is that the mean operation is a $\beta$-contraction. 
That is, given a transition matrix $P$, the operator $G$ defined by 
\begin{equation}\label{Qcontract}
GQ(x,a) = r(x,a) + \beta \mathbb E_{x,a} \Big[ \max_{a'\in\mathcal A}Q(\hat x,a') \Big]\,	
\end{equation}
is a $\beta$-contraction with respect to the supremum norm. I.e. for all $Q= (Q(s,a) : s \in \mathcal S, a \in \mathcal A)$ and $Q'= (Q'(s,a) : s \in \mathcal S, a \in \mathcal A)$  it holds that
\[
|| GQ - GQ'||_\infty \leq \beta || Q - Q' ||_\infty \, .
\]
(See Lemma \ref{lem:contraction} in the appendix.) We let $G_t$ to be the operator above \eqref{Qcontract} defined by transition matrix $P^{(t)}$.

\section{Adabiatic Theorem}\label{sec:adabiatic}

Our main result given here is Theorem \ref{thrm:adiabatic}. 
The result determines how, for a time inhomogeneous Markov chain, the rate of change in the sequence transition matrices $\{ P^{(t)} : t=0,...,T\}$ determines the closeness to the stationary distribution of a time-homogeneous Markov chain with transition matrix $P^{(T)}$.
To prove Theorem \ref{thrm:adiabatic}, we first require a supporting result, namely, Proposition \ref{prop:ergprop}, as well as, some standard lemmas given in the appendix. 
We then discuss the impact of mixing times and changes in the transition matrix on the sum of discounted rewards.

Proposition \ref{prop:ergprop} determines key properties of the coefficient of ergodicity, $\rho(P)$, defined by \eqref{eqn:ergodic}. The proposition collects together a number of results given by \cite{seneta1988perturbation,seneta2006non} and reviewed by \cite{ipsen2011ergodicity}. For completeness, a proof is given in the appendix.

 \begin{proposition}\label{prop:ergprop}
For two transition Matrices $P$ and $\tilde P$ and probability distributions $\lambda$ and $\mu$\\
a) 
\[
\rho_2 \leq \rho(P)
 \]	
where $\rho_2$ is the modulus of the 2nd largest eigenvalue of $P$.

\noindent b) 
\[
\rho(P\tilde P) \leq \rho(P) \rho(\tilde P)\, .
\]

\noindent c) 
\[
|| \lambda P - \mu P || \leq \rho(P) || \lambda - \mu || \,.
\]
\noindent d)
\begin{align*}
\rho(P)
&
= 
\max_{x_1,x_2}
|| P_{x_1,\cdot} - P_{x_2,\cdot}||\, .
\\
&
=
1 -
\min_{x_1,x_2} \sum_{y \in \mathcal X} \min \left\{ P_{x_1,y} , P_{x_2,y} \right\}	\, .
\end{align*}
\noindent e) 
\[
	|| \pi - \tilde \pi || \leq \frac{1}{1- \rho(P)}  \| P -\tilde P\|  \, .
\]
\end{proposition}

%
%
%
%
%
%
%
Part a) above shows that the coefficient of ergodicity is a valid proxy for the $2$nd largest eigenvalue, which determines the rate of convergence to equilibirum. Parts b) can be used to obtain linear convergence to equilibrium. Part d) gives a definition of the coefficient of ergodicity which can be directly calculated.  Parts c) and e) will be useful in proofs. 

The following gives the main result of this section. 

\begin{theorem}[Adiabatic Theorem] \label{thrm:adiabatic}
Let $P,P^{(1)},...,P^{(T)}$ be irreducible transition matrices. Let $\pi^{(T)}$ the stationary distribution of $P^{(T)}$ and $\rho(T)$ the coefficient of ergodicity of $P^{(T)}$. Let 
 $\mu$ and $\lambda$ be two probability distributions. The following holds:\\
a)
\[
|| \lambda P^{(1)}...P^{(T)} - \mu P^{T} ||
\leq
 || \lambda - \mu || \rho(P)^T 
+
\sum_{t=1}^T
		|| P^{(t)} - P || \rho(P)^{T-t}\, .
\]
b) If 
\begin{equation}\label{eqn:P_Conv}
	|| P^{(t)} - P^{(t-1)}|| \xrightarrow[t\rightarrow 0]{} 0\,  
\end{equation}
and 
\[
\limsup_{T\rightarrow\infty} \rho(T) < 1
\]
then 
\begin{equation}\label{eqn:PtoPi}
|| \lambda P^{(1)}...P^{(T)} - \pi^{(T)} || 
\xrightarrow[t\rightarrow \infty]{} 0 \, .	
\end{equation}
c) If 
\[
 ||
	P^{(t)} - P^{(t-1)} 
 ||
\leq \phi_t
\]
for a positive decreasing sequence $\phi_t$, then 
\[
|| \lambda P^{(1)}...P^{(T)} - \pi^{(T)} || 
\leq
\phi_{T/2}
\frac{\rho(T)}{(1-\rho(T))^2}
+
\frac{
\rho(T)^{T/2+1}
}{
1-\rho(T)
}
\sum_{t=1}^{T/2}
\phi_t
+
\|
\lambda - \pi^{(T)}
\|
\rho(T)^T \, .
\]
\end{theorem}

Before giving a proof, let's briefly interpret the three parts of the above result.
Note that the probability distribution at time $T$ of a time-homogeneous Markov chain with initial distribution $\mu$ and transition probabilities $P$ is $\mu P^T$, while for a time-inhomogeneous chain with initial distribution $\lambda$ and transition probabilities $P^{(1)},...,P^{(T)}$ it is $\lambda P^{(1)}...P^{(T)}$. 
Thus part a) determines how close these marginal distributions are at time $T$ give the distance in total variation between $P$ and $P^{(t)}$ and between $\mu$ and $\lambda$. 

Part b) gives a simple yet general adiabatic result which concerns the closeness to stationarity under change. 
We show that provided the coefficient of ergodicity is eventually less than $1$, then we need the change in the components of $P^{(t)}$ to go to zero for convergence of the empirical distribution at time $t$ to go to $\pi^{(t)}$.
Notice, importantly, this does not require convergence of $P^{(t)}$.
In other words, the sequence $P^{(t)}$ can vary over the set of irreducible matrices with $\rho(t)< \rho < 1$. So long as the change in $P^{(t)}$ is vanishingly small, then the current marginal distribution is close to the stationary distribution of the current chain.
Conversely, it is clear that if the condition \eqref{eqn:P_Conv} does not converge to zero then, in general, we cannot expect the result \eqref{eqn:PtoPi} to hold. 
Of course, if is possible for two transition matrices, to have the same stationary distribution, therefore the full converse cannot ever hold. 
Nonetheless, it should be clear that \eqref{eqn:P_Conv} provides a reasonably minimal necessary condition for \eqref{eqn:PtoPi} to hold.

Part c) improves on part b) by more accurately establishing rates of convergence. It should be clear from the statement of part c) that if $\phi_t= t^{-\eta}$ and $\rho(T)< \rho <1$  then the rate convergence to equilibrium is of the order $O(T^{-\eta})$. Another, perhaps, more important point is that the proof can be used to trade-off the rate of convergence exploited by the sequence $P^{(t)}$ and the exploration suggested by the coefficient of ergodicity $\rho$. 

\Beginproof
\noindent a) Recall, from Proposition \ref{prop:ergprop}, that 
\[
\|\lambda' P - \mu' P  \| \leq \rho(P) \| \lambda' - \mu' \|\,,
\]
and it follows from the definition of the total variation distance that 
\[
\| \lambda' (P'-P) \|
\leq 
\max_{x\in\mathcal X}\|  P'_{x \cdot} - P_{x \cdot}     \| = || P' - P ||\, .
\]
Applying the triangle inequality and the two inequalities above gives
\begin{align*}
  \| 
	\lambda P^{(1)} ...P^{(T)} - \mu P^T 
\|
&
\leq 
\|
\lambda P^{(1)} ... P^{(T-1)} (P^{(T)} -P)  
\|
+
\| \lambda P^{(1)} ... P^{(T-1)} P - \mu P^{T-1} P \|
\\
&
\leq 
\|
P^{(T)} -P
\|
+
\rho(P)
\|
\lambda P^{(1)} ... P^{(T-1)} - \mu P^{T-1}  
\|\, .
\end{align*}
By repeatedly iterating the above inequality, we have that 
\[
\| 
\lambda P^{(1)} ... P^{(T)}  - \mu P^T
\|
\leq 
\|
\lambda - \mu
\|
\rho(P)^T
+
\sum_{t=1}^T 
\|
P^{(t)} 
-
P
\|
\rho(P)^{T-t}
\]
as required.
\medskip

\noindent b) If we let $P=P^{(T)}$ and $\mu = \pi^{(T)}$, where $\pi^{(T)}$ is the stationary distribution of $P^{(T)}$, then part a) gives that 
\begin{equation}\label{eqn:express}
\|
\lambda P^{(1)} ... P^{(T)} - \pi^{(T)}
\|
\leq 
\|
\lambda - \pi^{(T)} 
\|
\rho (T) ^T
+
\sum_{t=1}^T 	\|
		P^{(t)}  - P^{(T)}
	\|
	\rho(T)^{T-t}\, .
\end{equation}
By the triangle inequality we have that 
\[
\|
	P^{(t)}- P^{(T)} 
\|
\leq 
	\sum_{s=t+1}^T 
\|
	P^{(s)} - P^{(s-1)}
\|
\,
.
\]
Applying this gives
\begin{align}
  \|
		\lambda P^{(1)}...P^{(T)} - \pi^{(T)} 
	\|
&
\leq 
\|
	\lambda - \pi^{(T)}
\|
\rho(T)^T
+
\sum_{t=1}^T \sum_{s=t+1}^T
		\|
			P^{(s)} 
			-
			Px^{(s-1)} 
		\|
\rho(T)^{T-t}
\notag
\\
&
=
\|
\lambda
- \pi^{(T)} 
\|
\rho(T)^T
+
\sum_{s=1}^T 
\|
P^{(s)} 
-
P^{(s-1)}
\|
\rho(T)^T 
\frac{\rho(T)^{-s} - 1}{\rho(T)^{-1} -1} \, .
\label{eq:Prop_Bound}
\end{align}
In the equality above we reorder the double summation and sum the resulting geometric series.

Since we assume
\[
\limsup_{T\rightarrow \infty} \rho(T)< 1 ,
\]
there exist $\rho <1$ and $T_0$ such that for all $T>T_0$, it holds that $\rho(T)< \rho < 1$.
For such values of $T$, the above expression \eqref{eqn:express} becomes 
\[
\|
\lambda P^{(1)} ... P^{(T)} - \pi^{(T)}
\|
\leq 
\|
\lambda
- \pi^{(T)} 
\|
\rho^T
+
\sum_{s=1}^T 
\|
P^{(s)} 
-
P^{(s-1)}
\|
\rho^T 
\frac{\rho^{-s} - 1}{\rho^{-1} -1} \, .
\]
Because 
\[
\| P^{(s)} - P^{(s-1)} \| \xrightarrow[s \rightarrow \infty]{} 0 \, ,
\]
the remainder of the proof of part b) follows by a Dominated Convergence Theorem argument. Specifically, take $\eta >0$ there exists $s_\eta$ such that for all $s \geq s_{\eta}$, 
\[
\|
	P^{(s)}- P^{(s-1)}
\|
\leq \eta 
\]
and of course for all $s \leq s_{\eta}$, $\max_{x\in\mathcal X} 
\|
	P^{(s)}_{x \cdot} - P^{(s-1)}_{x \cdot} 	
\|
\leq 1 $. 
Applying this to the summation in \eqref{eq:Prop_Bound} gives
\begin{align*}
\sum_{s=1}^T 
	\| 
		P^{(s)}- P^{(s-1)}
	\|
	\rho^{T-s} 
\leq 
\eta \sum_{s=s_{\eta}}^T \rho^{T-s} 
+
\sum_{s=1}^{s_{\eta}-1} 
	 \rho^{T-s} 
\leq 
\frac{\eta}{1-\rho} + \rho^T \sum_{s=1}^{s_{\eta}-1} \rho^{-s} 	\, .
\end{align*}
Therefore, applying the above bound to \eqref{eq:Prop_Bound} gives
\[
\limsup_{T\rightarrow \infty} \;\;
\|
\lambda P^{(1)}...P^{(T)} - \pi^{(T)}
\|
\leq 
\frac{\eta}{1-\rho} \, .
\]
Since $\eta$ can be made arbitrarily small the result for part b) holds.

\medskip

\noindent c) 
We now perform a closer analysis of the bound \eqref{eq:Prop_Bound}. 
We assume that 
\[
\|
P^{(t)} - P^{(t-1)} 
\|
\leq 
\phi_t \, .
\]
Applying this bound to the sum in \eqref{eq:Prop_Bound} gives
\begin{align*}
  \sum_{t=1}^T 
			\|
				P^{(t)} - P^{(t-1)}
			\|
\rho(T)^{T-t}
&\leq 
 \sum_{t=1}^T {\rho(T)^{T-t}} \phi_t
\\
&
=
 \sum_{t=T/2+1}^{T} {\rho(T)^{T-t}} \phi_t
+
 \sum_{t=1}^{T/2} {\rho(T)^{T-t}} \phi_t
\\
&
\leq 
\phi_{T/2} \sum_{t=T/2+1}^T {\rho(T)^{T-t}}
+
\rho(T)^{T/2} \sum_{t=1}^{T/2}
\phi_t
 \\
&
\leq 
\frac{\phi_{T/2}}{
1-\rho(T)
}
+
\rho(T)^{T/2} \sum_{t=1}^{T/2}
\phi_t \, .
\end{align*}
Now substituting this back into the bound \eqref{eq:Prop_Bound} gives
\begin{align*}
\| 
\lambda P^{(1)} ... P^{(T)}  - \mu P^T
\|
&
\leq 
\|
\lambda - \mu
\|
\rho(T)^T
+
\frac{1}{\rho(T)^{-1} -1}
\left[
\frac{\phi_{T/2}}{
1-\rho(T)
}
+
 \rho(T)^{T/2} \sum_{t=1}^{T/2}
\phi_t
\right]
\\
&
\leq 
\phi_{T/2}
\frac{\rho(T)}{(1-\rho(T))^2}
+
\frac{
\rho(T)^{T/2+1}
}{
1-\rho(T)
}
\sum_{t=1}^{T/2}
\phi_t
+
\|
\lambda - \pi^{(T)}
\|
\rho(T)^T
\end{align*}
as required.
\Endproof

\subsection{Discounted Reward Processes}

In this subsection, we focus on the sensitivity of the cumulative rewards to changes in the transition matrix. We recall the definition of the reward function $R(x)=R(x;P)$ from \eqref{reward_function}.

The following lemma is a consequence of Theorem \ref{thrm:adiabatic}. It shows that the reward function $R(x;P)$ and $Q$-function $Q(s,a;P)$ is Lipschitz continuous in $P$.
\begin{lemma}\label{RLem}
For a discounted program\\ a)
\[
\|
R(\cdot,P)
-
R(\cdot, \tilde P) \|_\infty 
\leq \frac{\beta r_{\max}}{(1-\beta)^2}
\| P- \tilde P\|
\]
b)
\[
\| Q(\cdot,\cdot; P) - Q(\cdot, \cdot ; \tilde P)\|_\infty
\leq 
\frac{\beta r_{\max}}{(1-\beta)^2}
\| P- \tilde P\|
\]
\end{lemma}

The following holds as a consequence of the above. It shows that the rewards of Markov chain (and thus a discounted program) can be expressed interms of a Markov chain with ergodicity coefficient strictly less than $1$.
\begin{lemma}\label{DiscRho}
	If $\hat x_t$ is a time homogeneous Markov chain then for $\tilde \beta \in ( \beta , 1)$ there exists a positive recurrent time-homogenous Markov chain $\tilde x_t$ whose transition matrix, $\tilde P$, satisfies
\[
\rho(\tilde P) < \frac{\beta}{\hat \beta} <1
\qquad \text{
and}
\qquad
R(x;\tilde P)
=
\frac{1-\beta}{1-\hat \beta}
R(x; P) 
\,.
\]
\end{lemma}
The result ensures that fast mixing can be achieved uniformly across all discounted Markov decision processes, which is important for a temporal difference learning algorithm to converge quickly,


\section{Asynchronous Stochastic Approximation} \label{sec:async}

We apply our adiabatic result to asynchronous stochastic approximation. We consider an asynchronous approximation problem where the target fixed point is changing in a time dependent way and so is the time-inhomogeneous Markov chain that determines which components are updated. Because of the time varying setting, we need to be careful to account for dependence on ergodicity coefficients and the minimum stationary probability. 

%
%
 At each time $t$ we are given an operator $R \mapsto F_tR$, where $F_t : \mathbb R^n \rightarrow \mathbb R^n$ is a $\beta$-contraction with respect to the supremum norm. We assume that $||F_t R||_\infty \leq \beta || R ||_\infty + F_{\max} $ for some positive constant $F_{\max}$. (We will shortly see that this property holds in the case of TD-learning.) We focus on the task of tracking $R^\star_t$ a fixed point 
\begin{equation}\label{eq:Fix}
F_{t}R^\star_t= R^\star_t.	
\end{equation}


Here we suppose that $(\hat x(t) : t \in \mathbb Z_+)$ is a time-homogeneous Markov chain with irreducible transition matrix $P^{(t)}$ at time $t$. We suppose that the coefficient of ergodicity is bounded above:
\[
\sup_{t\in\mathbb Z_+}\rho(P^{(t)}) < \rho \, 
\]
for some $\rho>0$. Recall that by Lemma \ref{DiscRho}, any discounted program can be simulated by a Markov chain satisfying this property. We  let $\pi^{(t)} = (\pi^{(t)}(x) : x \in\mathcal X)$ be the stationary distribution of $P^{(t)}$. We assume that 
\[
\pi^{(t)}_{\min} \geq \frac{C_{\pi}}{t^{\gamma_{\pi}}} 
\]
We assume that the sequence $(F_t : t\in\mathbb Z_+)$ is independent of the Markov chain $(\hat x(t):t\in\mathbb Z_+)$. In this sense we focus on tracking the fixed point as it changes over time rather than influencing its location.



We consider a stochastic approximation algorithm which at time $t$ maintains a vector $R_t = (R_t(x) : x \in\mathcal X)$.
We take $R_0(x)=0$ for $x\in\mathcal X$ and we update $R_t$ according to the rule
\begin{equation} \label{SA:update}
R_{t+1}(x) 
=
R_t(x)
+ 
\alpha_t 
\left[
F_t R_t(x) - R_t(x) + \epsilon_t
\right]
\qquad\text{ for }x = \hat x(t)	
\end{equation}
and $R_{t+1}(x) = R_t(x)$ if $x\neq \hat x(t)$. Here $\epsilon_t$ is a bounded martingale difference sequence with respect to filtration $\mathcal F_t$ generated by the past states $\hat x_s$, $s\leq t$. 
We assume that $\alpha_t$ is a power function that is
\[
\alpha_t= \frac{C_\alpha }{t^{\gamma_\alpha}} \, 
\]
for constant $C_{\alpha} \in (0,1)$ and for $\gamma_\alpha \in (0,1)$. 
We make the assumption that
\[
\gamma_\alpha + \gamma_{\pi} < 1.
\]
Also we assume that $P^{(t)}$ is both bounded above as follows
\begin{equation}\label{RP}
|| P^{(t+1)} - P^{(t)}||
\leq \frac{C_P}{t^{\gamma_P}} \, 	
\end{equation}
for positive constants $C_P$ and $\gamma_P$.
We assume that $F_t$ is dependent on $P_t$ in that $R_t^\star$ is Lipschitz in $P_t$. That is
\[
|| R^\star_{t+1} - R^\star_t ||_\infty \leq
K || P^{(t+1)} - P^{(t)} || \,.
\]
We recall Lemma \ref{RLem} for justification of this Lipschitz assumption in the context of dynamic programming.

\begin{theorem}\label{thrm:asych}
The distance between  the fixed point \eqref{eq:Fix} and the stochastic approximation \eqref{SA:update}, as described above, obeys the following bound with probability greater that $1-\delta$:
\begin{align}
	\| R_{T+1}(x) - R_{T+1}^\star(x) \|_\infty
\leq 
&
2 R_{\max} 
e^{(1-\beta)\tau} 
\exp\left\{ 
-
(T^{1-\gamma_\alpha-\gamma_\pi}-1)/(1-\gamma_\alpha-\gamma_\pi)
\right\}
\label{ada1}
\\
&
+
\frac{{D_a} }{1-\beta}
\sqrt{\tau \log \left(
\frac{2T \tau }{\delta}
\right)} 
\frac{1}{T^{(\gamma_\alpha-3\gamma_{\pi})/2}}
\label{ada2}
\\
&
+
\frac{{D_{b}}K}{(1-\beta)}
 \frac{1}{T^{\gamma_P-\gamma_\alpha - \gamma_{\pi}}}
\label{ada3}
\\
&
+
\frac{2 R_{\max}D_{b'}}{(1-\rho)^2(1-\beta)}\frac{ 1}{T^{\gamma_P-\gamma_{\pi}}}
+
\frac{8 R_{\max}}{(1-\rho)^2}
\frac{\log T}{T^3}
+
\frac{2R_{\max}}{T^3}\,,
\label{ada4}
\end{align}
where  $\tau := 4 \frac{\log T }{|\log \rho|}$, and $D_a$, $D_{b}$, $D_{b'}$ are constants depending only on $\gamma_P$, $C_P$, $\gamma_\alpha$, $C_\alpha$, $\gamma_\pi$, $C_\pi$.
\end{theorem}

Before proceeding with a proof we interpret the terms in Theorem \ref{thrm:asych}.  
Like with stochastic gradient descent, the term \eqref{ada1} corresponds to the exponential rate that we forget the initial condition. The term \eqref{ada2} accounts for the mixing/adiabatic time of the Markov chain. The bound requires $2 \gamma_\alpha > 3 \gamma_\pi$ for convergence. A conjecture is that the dependence should be $\gamma_\alpha >  \gamma_\pi$. In either case, we require step sizes to converge at a faster rate than the rate that we avoid states in the chain. 
The term \eqref{ada3} is the most important  term. We see that if
\[
\gamma_P > \gamma_\alpha + \gamma_\pi
\]
then we can track the fixed point solution. Thus the stochastic approximation scheme is adiabatic. However, if $\gamma_P < \gamma_\alpha + \gamma_\pi$ then we do not expect the stochastic approximation scheme to converge on the current fixed point, and thus is diabatic. 
The term \eqref{ada4} is dominated by earlier terms does not determine our rate of convergence; however, it does include the dependencies on the coefficient of ergodicity. Further we note that the $O(T^{-3})$ can be replaced with $O(T^{-n})$ for arbitrary $n$. Finally we note that if $\gamma_P-\gamma_\alpha-\gamma_\pi > (\gamma_\alpha-3\gamma_\pi)/2$ then the order of the convergence rate is the same as for the time-homogeneous stationary learning problem typically considered in asynchronous stochastic approximation.

A number of supporting lemmas are required in the proof of Theorem \ref{thrm:asych}, specifically, Lemmas \ref{lem:zcomp},  \ref{lemma:alpha_sum}, \ref{prodbound}, \ref{Lem:Atoa}, \ref{zboundLem}, \ref{AHbound} and \ref{lem7}.  These are stated immediately. after the proof of Theorem  \ref{thrm:asych} in Section \ref{sec:thrmlem}, and proofs are given in the appendix in Section \ref{app:4}.

\Beginproof[Proof of Theorem \ref{thrm:asych}]

	We can rewrite the update \eqref{SA:update} as 
\begin{align*}
R_{t+1}(x) 
&=
R_{t}(x)
+
\alpha_t \mathbb I [\hat x_t =x ] 
\Big[
F_tR_t(x) - R_t(x) + \epsilon_t
\Big]\,.
\end{align*}
Using the fact that $F_t R_t^\star(x) = R_t^\star(x)$ and adding \& subtracting terms, the above can be rewritten as
\begin{align}
  R_{t+1}(x) - R^\star_{t+1}(x)
=
&
(1-\alpha_t \pi^{(t)}(x))
\left[ 
  R_{t}(x) - R^\star_{t}(x)
\right]
\notag
\\
&
+
\alpha_t \pi^{(t)}(x) \left[
F_t R_t(x) - F_t R^\star_t(x)
\right] 
\notag
\\
&
+ \left[ R_t^\star (x) - R_{t+1}^\star (x) \right]  
\notag
\\
&
+
\alpha_t
\left[
\mathbb P( \hat x_t = x |\mathcal F_{t-\tau})
-\pi^{(t)}(x)
\right]
\left[
F_tR_t(x) -R_t(x)
\right]
\notag
\\
&
+
\alpha_t\left[ 
\mathbb I [\hat x_t = x] 
-
\mathbb P(\hat x_t = x | \mathcal F_{t-\tau} )
\right]  \left[
F_t R_t(x) -  R_t(x)
\right] 
\notag
\\
&
+ \alpha_t \mathbb I [\hat x_t = x]  \epsilon_t \, 
\label{RstarBound}
\end{align}
given the expression above we define 
\begin{align*}
	c_t(x) & := \left[
F_t R_t(x) - F_t R^\star_t(x)
\right] 
\\ 
b_t(x) & := \left[ R_t^\star (x) - R_{t+1}^\star (x) \right]  
\\
b'_t(x) & := 
\left[
\mathbb P( \hat x_t = x |\mathcal F_{t-\tau})
-\pi^{(t)}(x)
\right]
\left[
F_tR_t(x) -R_t(x)
\right]
\\
\epsilon'_t(x)& := \left[ \mathbb I [\hat x_t = x] 
-
\mathbb P(\hat x_t = x | \mathcal F_{t-\tau} ) \right]\left[
F_t R_t(x) - F_t R^\star_t(x)
\right] \, .
\end{align*}
Thus the expression above, \eqref{RstarBound}, can be more compactly written as
\[
R_{t+1}(x) - R^\star_{t+1}(x)
=
(1-\alpha_t \pi^{(t)}(x)) 
[R_t(x) - R^\star_t(x)]
+\alpha_t \pi^{(t)} (x) c_t(x) + b_t(x) + \alpha_t b'_t(x) + \alpha_t \epsilon_t + \alpha_t \epsilon'_t(x) \, .
\]
In order words, $R_{t+1}(x)- R^\star_{t+1}(x)$ is a combination of one contraction term $c_t$, two bias terms $b_t(x)$ and $b'_t(x)$ and two martingale difference terms $\epsilon_t$ and $\epsilon'_t(x)$. (By assumption $\epsilon_t$ is bounded and, by Lemma \ref{Lem:Bound} ---proved in the appendix--- $R_t(x)$ is bounded and so $\epsilon'_t(x)$ is bounded. We let $\epsilon_{\max}$ upper bound of their sum.)
Ideally we would apply the contraction property to $c_t$ at this point and then take a supremum over $x$; however, martingale difference sequence considered is dependent on $x$ and thus we will not be able to apply concentration inequalities to the martingale difference sequence. To avoid this difficulty, we first expand the iterations of the expression above to in order to apply Azuma-Hoeffding Inequality and then seek to apply the contraction property.  

Expanding the recursion using Lemma \ref{lem:zcomp} gives:
\begin{align}
	 R_{T+1}(x)
- R^\star_{T+1}(x)
= 
&
[R_\tau(x) - R_\tau^\star(x)]
\prod_{t=\tau}^T 
\left(
1 - \alpha_t \pi^{(t)}(x)
\right)
\notag\\
& 
+
\sum_{t=\tau}^T 
\left[
	\alpha_t \pi^{(t)}(x)c_t(x) 
	+
	b_t(x)
	+
	\alpha_t b'_t(x)
\right]
\prod_{s=t+1}^T 
(1- \alpha_t \pi^{(t)}(x))
\notag
\\
& 
+
\sum_{t=\tau}^T 
\alpha_t \left[
\epsilon_t 
+
\epsilon'_t(x)
\right]
\prod_{s=t+1}^T 
(1- \alpha_t \pi^{(t)}(x))
\label{Req}
\end{align}
We now bound each of the terms above to obtain our result.

We start with the martingale difference terms, by the Azuma-Hoeffding  Inequality given in Lemma \ref{AHbound} with probability at least $1-\delta$, it holds that for all $t\leq T$
\begin{align*}
\Bigg|
\sum_{s=\tau}^{t} 
\alpha_s
\left[
	\epsilon_s
	+
	\epsilon'_s(x)
\right]
\prod_{u=s+1}^{t} 
(1- \alpha_u \pi^{(u)}(x))
\Bigg|
&
\leq 
\sqrt{2\tau \sum_{s=1}^{t} 
\left[ \epsilon_{\max}^2 \alpha^2_{s}\prod_{u=s+1}^{t} 
(1- \alpha_u \pi_{\min}^{(u)})^2 \right] \log \left(\frac{2T \tau }{\delta}\right)}
\\
&
\leq 	
\frac{\epsilon_{\max} D_\alpha}{t^{(\gamma_\alpha-\gamma_{\pi})/2}}
\sqrt{ \tau\log \left( \frac{2T\tau}{\delta}\right)}\,
=: A_{t}.
\end{align*}
In the final inequality, we simplify the expression by applying Lemma \ref{prodbound} (noting that $(1-\alpha)\geq (1-\alpha)^2$). Here $D_\alpha$ is a constant only depending on $\gamma_\alpha$, $C_\alpha$, $\gamma_\pi$, $C_\pi$. 
Given $A_{t}$ as defined above and recalling Lemma \ref{Lem:Atoa}, we can define 
\[
a_t = \frac{A_t - A_{t-1}}{\alpha_t} + A_{t-1} \,
\]
and by Lemma \ref{Lem:Atoa},
\begin{equation} \label{aDef:Stuff}
	a_t \leq
\frac{\epsilon_{\max} D'}{t^{(\gamma_\alpha-\gamma_{\pi})/2}}
\sqrt{ \tau\log \left( \frac{2T\tau}{\delta}\right)} \, ,
\end{equation}
where $D'$ is a constant depending only on $\gamma_\alpha$, $C_\alpha$, $\gamma_\pi$ and $C_{\pi}$.

We can also bound the other terms $c_t(x)$, $b_t(x)$ and $b'_t(x)$. For $c_t(x)$ we apply the assumed contraction property that is
\begin{equation}\label{cBound}
	||c_t(\cdot)||_\infty=
\|
F_t R_t - F_t R^\star_t
\|_\infty 
\leq 
\beta 
\| R_t -  R^\star_t
\|_\infty
\end{equation}
For $b_t(x)$, we have by the assumed Lipschitz property (see also Lemma \ref{RLem}) that 
\begin{equation}\label{bstar}
||
R_t^\star  - R^\star_{t+1}
||_\infty
\leq 
K
 \| P^{(t)} -  P^{(t+1)}\| 
\leq 
K
\frac{C_P}{t^{\gamma_P}}
=: b^\star_t .	
\end{equation}
For $b'_t(x)$, we have by the Adiabatic Theorem 
 and more specifically by Lemma \ref{lem7} that:
\begin{align*}
\left\|
\mathbb P( \hat x_t = \cdot |\mathcal F_{t-\tau})
-\pi^{(t)}(\cdot)
\right\|
&
\leq 
\frac{1}{(1-\rho)^2}\frac{D_P}{t^{\gamma_P}} 
+
\frac{4}{(1-\rho)^2}
\frac{\log T}{T^4}
+
\frac{1}{T^4} \,,
\end{align*}
where $
\tau := 4 {\log T }/{|\log \rho|}.
$ 
Thus
\begin{align}
|| b'_t(\cdot) ||_\infty 
&
\leq   2R_{\max} 
\big\|
\mathbb P( \hat x_t = \cdot |\mathcal F_{t-\tau})
-\pi^{(t)}(\cdot)
\big\|
\notag
\\
&
\leq 
2R_{\max} 
\left[
\frac{1}{(1-\rho)^2}\frac{D_P}{t^{\gamma_P}} 
+
\frac{4}{(1-\rho)^2}
\frac{\log T}{T^4}
+
\frac{1}{T^4} 
\right]
=: b'^\star_t \, .	
\label{b_dash_star}
\end{align}

Given $a_t$ in \eqref{aDef:Stuff}, 
the bound for $c_t$ in \eqref{cBound} and the definitions of $b_t^\star$ in \eqref{bstar} and $b'^\star_t$ in \eqref{b_dash_star}, the equality \eqref{Req} now becomes the bound:
\begin{align*}
|R_{T+1}(x) - R^\star_{T+1}(x) |
\leq
&
|R_{\tau}(x) - R^\star_{\tau}(x)|
\prod_{t=\tau}^T 
\left(
1 - \alpha_t \pi^{(t)}(x) \right)
\\
&
+ \sum_{t=\tau}^T
\alpha_t \beta 
\| R_t -  R^\star_t
\|_\infty 
\prod_{s=t+1}^T 
(1- \alpha_t \pi^{(t)}(x))
\\
&+
\sum_{t=\tau}^T 
\left[
\alpha_t a_t
	+
	b_t^{\star}
	+
	\alpha_t b_t'^{\star}
\right]
\prod_{s=t+1}^T 
(1- \alpha_t \pi^{(t)}(x)) \,.
\end{align*}
Notice that we now have removed the martingale terms, and also applied the bounded terms with the Adiabatic Theorem (via Lemma \ref{lem7}). We now focus on re-introducing the $\beta$-contraction term. 

By Lemma \ref{zboundLem}
\[
|R_{T+1}(x) + R^\star_{T+1}(x)|
\leq z_{T+1}
\]
where $z_{t}$ obeys the recursion
\begin{align*}
z_{t+1} 
&
= (1-\alpha_t\pi^{(t)}(x)) z_t + \alpha_t \beta \pi^{(t)}(x) z_t 
+
\alpha_t a_t
+ 
b^\star_t + \alpha_t b'^{\star}_t \, .
\\
&
= 
(1-\alpha_t (1-\beta ) \pi^{(t)}(x))
	z_{t} 
+
\alpha_t a_t
+ 
b^\star_t + \alpha_t b'^{\star}_t 
\\
&
\leq 
(1-\alpha_t (1-\beta ) \pi^{(t)}_{\min})
	z_{t} 
+
\alpha_t a_t
+ 
b^\star_t + \alpha_t b'^{\star}_t  \, .
\end{align*}
for $t \geq \tau$ 
and $z_\tau = \| R_{\tau}(x) - R_{\tau}^\star(x) \|_\infty$.
Thus expanding this recursion using Lemma \ref{lem:zcomp}, we have that
\begin{align}\label{expander2}
&
\| R_{T+1}(x) - R_{T+1}^\star(x) \|_\infty
\notag
\\
&
\leq 
\| R_{\tau}(x) - R_{\tau}^\star(x) \|_\infty
\prod_{t=\tau}^T (1- \alpha_t (1-\beta) \pi^{(t)}_{\min})
+
\sum_{t=\tau}^T	
\left( 
\alpha_t a_t + b^\star_t + \alpha_t b'^\star_t
\right)
\prod_{s=t+1}^T 
(1- \alpha_s (1-\beta) \pi^{(t)}_{\min})\, .
\end{align}
We bound the above terms for $a_t$, $b^\star_t$, $b'^\star_t$ by appling Lemma \ref{prodbound}. Specifically, recalling \eqref{aDef:Stuff},
\begin{align}\label{abound}
	\sum_{t=\tau}^T \alpha_t a_t 
\prod_{s=t+1}^T (1-\alpha_s (1-\beta ) \pi^{(t)}_{\min})
&
\leq 
{\epsilon_{\max} D'} 
\sqrt{\tau \log \left(
\frac{2T \tau }{\delta}
\right)} 
\sum_{t=\tau}^T \
\frac{1}{t^{(\gamma_\alpha-\gamma_{\pi})/2}}
\alpha_t
\prod_{s=t+1}^T (1-\alpha_s (1-\beta ) \pi^{(t)}_{\min})
\notag
\\
&
=
\frac{\epsilon_{\max}{D_a} }{1-\beta}
\sqrt{\tau \log \left(
\frac{2T \tau }{\delta}
\right)} 
\frac{1}{T^{(\gamma_\alpha-3\gamma_{\pi})/2}}
\end{align}
where $D_a$ is a constant depending on $\gamma_\alpha$, $C_\alpha$, $C_{\pi}$, and $\gamma_\pi$;
also, recalling \eqref{bstar} and again applying Lemma \ref{prodbound}
\begin{align}
	\sum_{t=\tau}^T 
	b^\star_t
\prod_{s=t+1}^T
(1-\alpha_s(1-\beta)\pi^{(t)}_{\min})
&
\leq 
{{C_P}K}
\sum_{t=\tau}^T 
 \frac{1}{t^{\gamma_P-\gamma_\alpha}}
\alpha_t
\prod_{s=t+1}^T (1-\alpha_s (1-\beta )\pi^{(s)}_{\min})
\notag
\\
&
\leq 
{D_{b} K}
 \frac{1}{T^{\gamma_P-\gamma_\alpha - \gamma_{\pi}}}
\label{b_bound}
\end{align}
where $D_{b}$ is a constant depending on $\gamma_\alpha$, $C_\alpha$, $\gamma_\pi$, $C_\pi$, $\gamma_P$ and $C_P$; 
and 
\begin{align}\label{b2bound}
&\sum_{t=\tau}^T 
	\alpha_t
b'^\star_t
\prod_{s=t+1}^T
(1-\alpha_s(1-\beta)\sigma)]
\notag
\\
&= 
2R_{\max} 
\sum_{t=\tau}^T
\left[
\frac{1}{(1-\rho)^2}\frac{D_P}{t^{\gamma_P}} 
+
\frac{4}{(1-\rho)^2}
\frac{\log T}{T^4}
+
\frac{1}{T^4} 
\right]\alpha_t
\prod_{s=t+1}^T 
(1- \alpha_s (1-\beta) \pi^{(s)}_{\min})
\notag
\\
&
\leq 
\frac{2 R_{\max}D_{b'}}{(1-\rho)^2(1-\beta)}\frac{ 1}{T^{\gamma_P-\gamma_{\pi}}}
+
\frac{8 R_{\max}}{(1-\rho)^2}
\frac{\log T}{T^3}
+
\frac{2R_{\max}}{T^3}\,.
\end{align}
where $D_{b'}$ depends on $\gamma_P$, $C_P$, $\gamma_\alpha$, $C_\alpha$, $\gamma_\pi$, and $C_\pi$.

Further we can bound the term corresponding to the forgetting the estimate at time $\tau$:
\begin{align}\label{initbound}
& 
\| R_{\tau}(x) - R_{\tau}^\star(x) \|_\infty
\prod_{t=\tau}^T (1- \alpha_t (1-\beta) \pi^{(t)}_{\min})
\notag
\\
\leq 
&
2 R_{\max} 
\exp \left\{
- (1-\beta) \sum_{t=\tau}^T \alpha_t \pi^{(t)}_{\min}
\right\}
\notag
\\
\leq 
&
2 R_{\max} 
e^{(1-\beta)\tau} 
\exp\left\{ 
(1-\beta)\sum_{t=1}^T \frac{C_{\alpha} C_{\pi} }{t^{\gamma_\alpha+ \gamma_\pi}}
\right\}
\notag
\\
\leq
& 
2 R_{\max} 
e^{(1-\beta)\tau} 
\exp\left\{ 
-C_{\alpha} C_{\pi}(1-\beta)
(T^{1-\gamma_\alpha-\gamma_\pi}-1)/(1-\gamma_\alpha-\gamma_\pi)
\right\}
\notag
\\
\leq 
&
2 R_{\max} 
e^{(1-\beta)\tau} 
\exp\left\{ 
-
(T^{1-\gamma_\alpha-\gamma_\pi}-1)/(1-\gamma_\alpha-\gamma_\pi)
\right\}\, .
\end{align}
Above we apply the bound $(1-z) \leq e^{-z}$; we add and subtract then bound the first $\tau$ terms in the summation; we apply Lemma \ref{lemma:alpha_sum}; and then simplify by observing that $C_\alpha, C_\pi$, and $(1-\beta)$ are all less than $1$.

Thus applying the above inequalities \eqref{abound}, \eqref{b_bound}, \eqref{b2bound} and \eqref{initbound} to \eqref{expander2} gives the bound:
\begin{align*}
	\| R_{T+1}(x) - R_{T+1}^\star(x) \|_\infty
\leq 
&
2 R_{\max} 
e^{(1-\beta)\tau} 
\exp\left\{ 
-
(T^{1-\gamma_\alpha-\gamma_\pi}-1)/(1-\gamma_\alpha-\gamma_\pi)
\right\}
\\
&
+
\frac{{D_a} }{1-\beta}
\sqrt{\tau \log \left(
\frac{2T \tau }{\delta}
\right)} 
\frac{1}{T^{(\gamma_\alpha-3\gamma_{\pi})/2}}
\\
&
+
\frac{{D_{b}}K}{(1-\beta)}
 \frac{1}{t^{\gamma_P-\gamma_\alpha - \gamma_{\pi}}}
\\
&
+
\frac{2 R_{\max}D_{b'}}{(1-\rho)^2(1-\beta)}\frac{ 1}{T^{\gamma_P-\gamma_{\pi}}}
+
\frac{8 R_{\max}}{(1-\rho)^2}
\frac{\log T}{T^3}
+
\frac{2R_{\max}}{T^3}\, ,
\end{align*}
as required.
\Endproof


\subsection{Lemmas for Theorem \ref{thrm:asych}}\label{sec:thrmlem}

We now list additional lemmas that are required for Theorem \ref{thrm:asych}. We only state these lemma below. We restate and then prove the lemmas in Section \ref{app:4} of the appendix.

Lemma \ref{lem:zcomp} is a standard expansion commonly used in stochastic approximation. 
\begin{lemma}\label{lem:zcomp}
Suppose that $z_n$ is a positive real valued sequence such that 
\[
z_{n+1} \leq z_n (1 - a_n) + c_n
\]	
then 
\[
z_{n+1} \leq z_0 \prod_{k=0}^n (1-a_k)  +  \sum_{j=0}^n c_j \prod_{k=j+1}^n (1 - a_k)\, .
\]
\end{lemma}

Lemma \ref{lemma:alpha_sum} is a well-known integral bound.
\begin{lemma}\label{lemma:alpha_sum}
	If we let $\alpha_t = t^{-\gamma}$, for $\gamma \in (0,\infty)$, we have  
	\[
	\frac{t^{1-\gamma} - s^{1-\gamma}}{ 1-\gamma }  
	\leq \sum_{n=s}^t  \frac{1}{n^{\gamma}} 
	\leq
	\frac{1}{s^{\gamma}} + \frac{t^{1-\gamma} - s^{1-\gamma} }{ 1-\gamma }  \, .
	\] 
where for $\gamma = 1$, we define
$\;
{(t^{1-\gamma} - 1)}/{(1-\gamma )}  := \log t
$.
\end{lemma}

The following lemma is a commonly used bound that combines Lemma \ref{lem:zcomp} and Lemma \ref{lemma:alpha_sum}.
\begin{lemma}\label{prodbound}
	For positive sequences $a_t$ and $b_t$ with $a_t \in(0,1)$ and $b_t$ decreasing
\[
\sum_{t=1}^T a_t b_t \prod_{s=t+1}^T (1-a_s) 
\leq 
b_{T/2} 
+
e^{-\sum_{t=T/2}^T a_t}
\sum_{t=1}^{T/2} a_t b_t
\]
Moreover if $a_t = \frac{C_a}{t^{\gamma_a}}$ and $b_t = \frac{C_b}{t^{\gamma_b}}$ for $\gamma_a \in(0,1)$ and $\gamma_b \geq 0$ then 
\[
\sum_{t=1}^T a_t b_t \prod_{s=t+1}^T (1-a_s) 
\leq 
\frac{D_{a,b}}{T^{\gamma_b}}
\, ,
\]
where $D_{a,b}$ is a constant depending on $C_a,C_b,\gamma_a$ and $\gamma_b$.
\end{lemma}

The following lemma is a straight-forward converse to Lemma \ref{lem:zcomp} and appears to be less commonly used. 
\begin{lemma}\label{Lem:Atoa}
For any sequence $A_T$ we can write $A_T$ as
\[
A_T 
=
\sum_{t=1}^T
a_t \alpha_t
\prod_{s=t+1}^T (1-\alpha_s)
\]
where 
\[
a_t = \frac{A_t - A_{t-1}}{\alpha_t} + A_{t-1}\,.
\]
Thus if $A_t = C_A/ t^{\gamma_A}$ and $\alpha_t = C_\alpha/ t^{\gamma_\alpha}$ for $\gamma_{\alpha} \in (0,1]$ then 
\[
a_t \leq \frac{D}{t^{\gamma_A}}
\]
where $D$ is a positive constant depending on $\gamma_{\alpha}$, $\gamma_A$, $C_{\alpha}$, $C_A$.
\end{lemma}

Lemma \ref{zboundLem}, below, is also a slightly less standard bound involving the expansion from Lemma \ref{lem:zcomp}.

\begin{lemma}\label{zboundLem}For positive sequence $z_t$ and $\alpha_t \in (0,1)$ if
\[
z_{t+1}
\leq 
z_0 \prod_{s=1}^t (1-\alpha_s) 
+
\sum_{s=1}^t \beta  z_s 
\prod_{u=s+1}^t (1-\alpha_u)
+
\sum_{s=1}^t c_s 
\prod_{u=s+1}^t (1-\alpha_u)
\]
then $z_t \leq \tilde z_t$ where $\tilde z_t$ solves the recursion
\[
\tilde z_{t+1} = (1-\alpha_t (1-\beta)) \tilde z_t + c_t
\]
with $\tilde z_0\geq  z_0$.
\end{lemma}

Lemma \ref{AHbound} is a shifted Azzuma-Heoffding bound which can be found in \cite{qu2020finite}. 

\begin{lemma}\label{AHbound}
	If $\epsilon_t$ is adapted with $\mathbb E[ \epsilon_t | \mathcal F_{t-\tau}]=0$ and $|\epsilon_t| \leq \epsilon_{\max}$ then with probability greater than $1-\delta$  it holds that, for all $t$ with $\tau \leq t \leq T$,
\[
\left| 
\sum_{s=\tau}^t 
	\alpha_s 
\epsilon_s 
\prod_{u=s+1}^t (1-\alpha_s \pi_s)
\right| 
\leq 
\sqrt{
2  \tau
\left[\sum_{s=1}^t \epsilon^2_{\max} \alpha_s^2
\prod_{u=s+1}^t 
(1-\alpha_s\pi_s)^2
\right]
 \log \bigg(\frac{2\tau T}{\delta}\bigg)
} \, .
\]
\end{lemma}

Lemma \ref{lem7} is a mixing time result that applies the adiabatic theorem in the context of asynchronous stochastisc approximation,.

\begin{lemma}\label{lem7} 
For $\tau = 8 \frac{\log T }{|\log \rho|}$ and $t$ such that $\tau \leq t \leq T$ it holds that 
\[
\|
\mathbb P( \hat x_t = \cdot | \mathcal F_{t-\tau})
-
\pi^{(t)}(\cdot)
\|
\leq 
\frac{1}{(1-\rho)^2}\frac{D_P}{t^{\gamma_P}} 
+
\frac{4}{(1-\rho)^2}
\frac{\log T}{T^4}
+
\frac{1}{T^4}\, ,
\]
where $D_P =C_P 2^{\gamma_P}$.
\end{lemma}

\begin{lemma}\label{Lem:Bound}
	The sequence $R_t$ defined in \eqref{SA:update} and $F_tR_t$
 are bounded in $t$.
\end{lemma}



\section{Application to Tabular Reinforcement Learning}

We now show how the above results apply in the context of temporal difference learning and also $Q$-learning.

\subsection{Temporal Difference Learning}\label{sec:TD}

We consider tabular temporal difference learning as described in Section \ref{TabTD}.
We assume the transition matrix evolves in time. We let  $P^{(t)}$ be the transition matrix of the $t$-th transition.
We assume
\[
|| P^{(t+1)} - P^{(t)}||
\leq \frac{C_P}{t^{\gamma_P}} \, 	.
\]
We then seek to evaluate the fixed point equation:
\[
R^\star_t(x) =r(x) + \beta  P^{(t)}R_t^\star( x) \, .
\]
Note that the operation $F_t$ such that 
$
F_tR (x) = r(x) + \beta P^{(t)}R(x)
$
is a $\beta$-contraction (see Lemma \ref{lem:contraction}) and by definition $F_tR^\star_t = R^\star_t$. By Lemma \ref{RLem}
\[
\|
R^\star_t
-
R^\star_{t+1}\|_\infty 
\leq \frac{\beta r_{\max}}{(1-\beta)^2}
\| P^{(t)} - P^{(t+1)}\| \leq 
\frac{\beta r_{\max}}{(1-\beta)^2}
\frac{C_P}{t^{\gamma_P}} 
=:
\frac{C_R}{t^{\gamma_R}} \,.
\]

We suppose that $(\hat x_t : t \in\mathbb Z_+)$ is a time inhomogeneous Markov chain with transition matrix $P^{(t)}$ at time $t$. We consider the tabular temporal difference update:
\begin{align*}
R_{t+1}(\hat x_t) =
R_{t}(\hat x_t) 	
+
 \alpha_t 
\left[
r(\hat x_t) 
+ \beta 
R_t(\hat x_{t+1}) - R_t(\hat x_t)
\right]
\end{align*}
and $R_{t+1}(x) = R_t(x)$ for all $x\neq \hat x_t$. As before we assume 
$
\alpha_t = \frac{C_{\alpha}}{t^{\gamma_\alpha}}
$
and $\pi^{(t)}$ the stationary distribution of $P^{(t)}$ satisfies 
$
\pi^{(t)}_{\min} \geq \frac{C_{\pi}}{t^{\gamma_{\pi}}}  \, .
$
Identifying the above terms with the statement of Theorem \ref{thrm:asych} gives the following result.

\begin{theorem}
\begin{align*}
  	\| R_{T+1}(x) - R_{T+1}^\star(x) \|_\infty
\leq 
&
2 R_{\max} 
e^{(1-\beta)\tau} 
\exp\left\{ 
-
(T^{1-\gamma_\alpha-\gamma_\pi}-1)/(1-\gamma_\alpha-\gamma_\pi)
\right\}
\\
&
+
\frac{{D_a} }{1-\beta}
\sqrt{\tau \log \left(
\frac{2T \tau }{\delta}
\right)} 
\frac{1}{T^{(\gamma_\alpha-3\gamma_{\pi})/2}}
\\
&
+
\frac{{D_{b}}\beta r_{\max}}{(1-\beta)^3}
 \frac{1}{T^{\gamma_P-\gamma_\alpha - \gamma_{\pi}}}
\\
&
+
\frac{2 R_{\max}D_{b'}}{(1-\rho)^2(1-\beta)}\frac{ 1}{T^{\gamma_P-\gamma_{\pi}}}
+
\frac{8 R_{\max}}{(1-\rho)^2}
\frac{\log T}{T^3}
+
\frac{2R_{\max}}{T^3}\,.
\end{align*}
\end{theorem}
\subsection{Q-Learning}
We now consider $Q$-learning which is a variant of the temporal difference learning. Here we consider the fixed point:
\[
Q^\star_t = G^{(t)} Q^\star_t
\]
where 
\[
G^{(t)}Q(s,a) = r(s,a) + \beta \mathbb E^{(t)} \left[ \max_{a' \in\mathcal A} Q(\hat s,a') \right] \, .
\]

We perform the $Q$-learning update with respect to the time-inhomogeneous Markov chain $(\hat s^{(t)}, \hat a^{(t)})$ with transition probabilities $P^{(t)}$ at time $t$. That is 
\[
Q_{t+1} (\hat s_{t}, \hat a_{t}) 
= 
Q_t(\hat s_{t}, \hat a_{t}) 
+
 \alpha_t 
\left[
r(\hat s_t,\hat a_t) 
+ \beta 
\max_{a\in\mathcal A} Q_t(\hat s_{t+1},a) - Q_t(\hat s_t, \hat a_t)
\right] \, .
\]
and $Q_{t+1}(s,a) = Q_t(s,a)$ for all $(s,a)$ such that $s\neq \hat s_t$ or $a \neq \hat a_t$. Then as above 
we assume 
$
\alpha_t = \frac{C_{\alpha}}{t^{\gamma_\alpha}}
$
and $\pi^{(t)}$ the stationary distribution of $P^{(t)}$ satisfies 
$
\pi^{(t)}_{\min} \geq \frac{C_{\pi}}{t^{\gamma_{\pi}}}  \, .
$
We assume
\[
|| P^{(t+1)} - P^{(t)}||
\leq \frac{C_P}{t^{\gamma_P}} \, 	.
\]
By Lemma \ref{RLem} we have that
\[
\| 
Q^\star_t - Q^\star_{t+1}
\|
\leq \frac{\beta r_{\max}}{(1-\beta)^2}|| P^{(t+1)} - P^{(t)} ||
\]
Identifying the above terms with the statement of Theorem \ref{thrm:asych}, we also can obtain the following analogous result for $Q$-learning:

\begin{theorem}
\begin{align*}
  	\| Q_{T+1}- Q_{T+1}^\star \|_\infty
\leq 
&
2 R_{\max} 
e^{(1-\beta)\tau} 
\exp\left\{ 
-
(T^{1-\gamma_\alpha-\gamma_\pi}-1)/(1-\gamma_\alpha-\gamma_\pi)
\right\}
\\
&
+
\frac{{D_a} }{1-\beta}
\sqrt{\tau \log \left(
\frac{2T \tau }{\delta}
\right)} 
\frac{1}{T^{(\gamma_\alpha-3\gamma_{\pi})/2}}
\\
&
+
\frac{{D_{b}}\beta r_{\max}}{(1-\beta)^3}
 \frac{1}{T^{\gamma_P-\gamma_\alpha - \gamma_{\pi}}}
\\
&
+
\frac{2 R_{\max}D_{b'}}{(1-\rho)^2(1-\beta)}\frac{ 1}{T^{\gamma_P-\gamma_{\pi}}}
+
\frac{8 R_{\max}}{(1-\rho)^2}
\frac{\log T}{T^3}
+
\frac{2R_{\max}}{T^3}\,.
\end{align*}
\end{theorem}

\section{Conclusions and Future Work.}

As discussed in the introduction, usually theoretical results assume that policy evaluation algorithms are training with respect to a fixed reference policy. Consequently mixing time assumptions have generally been made in advance. This leads to important results. However, in practice this is very rarely the case. To make even simple algorithms converge the reference policy of interest is usually changed in time.
 This work takes a more in depth look at the effect of mixing times and adiabatic properties which effect the ability of reinforcement learning algorithms to convergence on a target process as it changes in time. We prove a new mixing time result which could be of independent interest and with this we can highlight issues on the conditioning of the stationary distribution and effects that occur from the changes in the learning target. 

The results proven give a better indication of the robustness of stochastic approximation and temporal difference learning to changes in transitions in probability distribution. From this we can see that the key condition for adiabatic TD-learning is that $\gamma_P > \gamma_\alpha + \gamma_\pi$. I.e. the rate of change in $P$ is faster than the sum of the rate of change in the learning rate plus the rate that the least likely equilibrium state goes to zero. Similarly the condition $\gamma_P-\gamma_\alpha-\gamma_\pi > (\gamma_\alpha-3\gamma_\pi)/2$ is required for the convergence rates to be the same as the stationary asynchronous stochastic approximation scheme.

There are certainly a number of directions in which this work can be generalized and developed. In this paper we only consider the policy evaluation process, we separate the changes in the transition matrix $P$ from the updates in the temporal difference learning algorithm. It is of course possible to allow for changes in $P$ to depend on the convergence of the temporal difference learning algorithm. Such as in actor-critic algorithms. Such results may depend on the specific form of that the update in $P$ depends on the temporal difference method and techniques such as the use of belief states (which are required for a Markov description in the POMDP setting) likely necessary to form an analysis. This introduces technical difficulties which would complicate the analysis. Nonetheless one would anticipate similar conclusions to the results proved in this paper. 

In this paper we consider stochastic approximation and tabular temporal difference learning. However, it is also important to consider function approximation in reinforcement learning. The Adiabatic Theorem, Theorem \ref{thrm:adiabatic}, is applicable in the case of temporal difference learning with linear function approximation. The set of techniques required are somewhat different to those applied here. In particular, we need to use online convex optimization methodology rather than stochastic approximation bounds. 
Given the differences in techiques we leave this work on adiabatic bounds in online convex optimization and linear temporal difference learning as forthcoming work.

\bibliography{mybibfile}

\appendix

\section{Additional Proof for Section \ref{sec:model}}

The following lemma is standard.
\begin{lemma}\label{lem:contraction}
	For any transition matrix $P$, the operators $F$ and $G$ defined by 
\[
FR(x) = r(x) + \beta PR(x)\qquad 
\text{and}
\qquad 
GQ(x,a) = r(x,a) + \beta \mathbb E_{x,a} \Big[ \max_{a'\in\mathcal A}Q(\hat x,a') \Big]\,	
\]
are both a $\beta$-contraction with respect to $||\cdot ||_\infty$.
\end{lemma}
\Beginproof
a)
First we prove the result for $F$. For $R=(R(x) : x \in\mathcal X)$ and $R'(x) = (R'(x) : x \in\mathcal X)$ it holds that
	\[
\big\| 
FR - F\tilde R
\big\|_\infty
=
\beta \sup_x \Big| \sum_{y} P_{xy}R(y)- P_{xy}R'(y) \Big|
\leq \beta  \sup_x \sum_{y} P_{xy} || R-R'||_\infty
=
\|
R-R'
\|_\infty\,. 
\]
For $G$ the proof is similar
\[
GQ(x,a) - G\tilde Q({x,a})
=
\beta 
\mathbb E_{x,a} [  \max_{a'} G Q(\hat{X},a') - 
\max_{a''} G \tilde Q(\hat{X},a'') ]
\]
Thus 
\begin{equation}\label{eqd}
	|| GQ - G\tilde Q||_{\infty}
\leq 
\beta 
  \max_{\hat{x}} | \max_{a'} G Q(\hat{x},a')- \max_{a''}  G \tilde Q(\hat{x},a'') |
 \leq 
  \beta 
  || Q_V - Q_R ||_{\infty}\, .
\end{equation}
In the last equality above, we use the fact that
\[
| \max_{a'} G Q(x,a')
- \max_{a'} G \tilde Q(x,a') |
\leq
\max_{a} | Q(x,a) - Q(x,a)|.
\]
\Endproof

\section{Additional proofs for Section \ref{sec:adabiatic}}

Proposition \ref{prop:ergprop} is a collection of existing results on the coefficient of ergodicity. These can be found in the book, \cite{seneta2006non}. Also the more recent survey of \cite{ipsen2011ergodicity} contains these results. The proof of part d) is taken from \cite{seneta1988perturbation}.

 \begin{Repeatproposition}{prop:ergprop}
For any two irreducible transition Matrices $P$ and $\tilde P$ and probability distributions $\lambda$ and $\mu$\\
a) 
\[
\rho_2 \leq \rho(P)
 \]	
where $\rho_2$ is the modulus of the 2nd largest eigenvalue of $P$.

\noindent b) 
\[
\rho(P\tilde P) \leq \rho(P) \rho(\tilde P)\, .
\]

\noindent c) 
\[
|| \lambda P - \mu P || \leq \rho(P) || \lambda - \mu || \,.
\]
\noindent d)
\begin{align*}
\rho(P)
&
= 
\max_{x_1,x_2}
|| P_{x_1,\cdot} - P_{x_2,\cdot}||\, .
\\
&
=
1 -
\min_{x_1,x_2} \sum_{y \in \mathcal X} \min \left\{ P_{x_1,y} , P_{x_2,y} \right\}	\, .
\end{align*}
\noindent e) 
\[
	|| \pi - \tilde \pi || \leq \frac{1}{1- \rho(P)}  \max_{x\in\mathcal X} \left\| P_{x, \cdot} -\tilde P_{x, \cdot} \right\|  \, .
\]
\end{Repeatproposition}

\Beginproof
	a) For the largest eigenvalue of $1$, the vector of ones $\bm 1 = (1 : x \in \mathcal X)$ is the unique right-eigenvector of the matrix $P$ and the stationary distribution $\pi$ is the unique left-eigenvector. (The uniqueness follows by irreducibility and the Perron--Frobenius Theorem.) 
Any left-eigenvector $\bm v$ with eigenvalue $\lambda$ with $|\lambda | < 1$ is orthogonal to $\bm 1$:
\[
\lambda (\bm v^\top \bm 1)  = \bm v^\top P 1= (\bm v^\top \bm 1) \qquad\text{thus}\qquad  \bm v^\top \bm 1 =0 \,.
\]
Consequently, normalizing $\bm v$ so that $||\bm v||=1$ gives:
\[
	\rho(P)
= 
\sup_{
\substack{ w: ||  w|| =1\\  w \cdot  \bm 1 = 0}  }
 || w P ||
\geq 
 || \bm v P || =| \lambda | \, .
\]
b) Note that we can equivalently write $\rho(P)$ as 
\[
\rho(P) = \sup_{\substack{v : v\neq 0\\v \cdot 1 = 0}}
\frac{|| vP || }{||v||} \, .
\]
Under this,
\begin{align*}
  \rho(P\tilde P)
= 
 \sup_{\substack{v : v\neq 0\\v \cdot 1 = 0}}
\frac{|| vP\tilde P || }{||v||} 
=
 \sup_{\substack{v : v\neq 0\\
vP  \neq 0
\\
v \cdot 1 = 0}}
\frac{|| vP\tilde P || }{||vP||} 
\frac{|| vP|| }{||v||} 
\leq 
 \sup_{\substack{u : u\neq 0\\ u \cdot 1 = 0}}
\frac{|| u\tilde P || }{||u||} 
 \sup_{\substack{v : v\neq 0\\ v \cdot 1 = 0}}
\frac{|| vP|| }{||v||} \,.
\end{align*}
(Note the supremum being positive cannot be attained by a vector for which $vP =0$.)

\noindent c) The bound trivially holds for $\lambda= \mu$. Since $(\lambda- \mu) \cdot 1 = 0$
\[
\frac{|| \lambda P - \mu P || }{|| \lambda - \mu ||}
\leq 
 \sup_{\substack{v : v\neq 0\\v \cdot 1 = 0}}
\frac{|| vP || }{||v||}
= \rho(P)\,.
\]
This following from the definition. 

\noindent d) Clearly
\[
\rho(P) \geq \Big\| \frac{(\bm e_x - \bm e_y) }{2} P \Big\| = \sum_{x'} | P_{xx'} - P_{yx'}| \, .
\]
Now maximizing over $x$ and $y$ gives a lower-bound.

For the inequality in the other direction.
We apply the decomposition from Lemma \ref{AuxLem} (which we state and prove immediately after this proof). That is $v$ such that $||v||_1=1$ and $v^\top \bm 1 = 0$ can be expressed as $v = \sum_{x\neq y} u_{xy} (\bm e_x- \bm e_y)/2$ where $\sum_{x\neq y} u_{xy} = 1$. Thus
\[
|| vP ||_1
\leq 
\sum_{x\neq y} u_{xy} \Big\|\frac{(\bm e_x - \bm e_y) }{2} P \Big\|
\leq \frac{1}{2}\max_{x,y} 
\sum_{x'} | P_{xx'} - P_{yx'}| \, .
\]

The second inequality in part d) follows since for any two probability distributions $p$ and $q$ it holds that 
\[|p_x-q_x|= p_x\vee q_x - p_x\wedge q_x\qquad\text{ and }\qquad p_x+ q_x = p_x\vee p_x + p_x\wedge q_x \, .
\]
Thus 
\[
\frac{1}{2} |p_x - p_y | = \frac{p_x + p_y}{2} - p_x \wedge p_y \, . 
\]
Summing gives $|| p - q || = 1-\sum_x p_x \wedge q_x$. Applying this to the rows in $P$ gives the result.

e)
We let 
\[
 v = {\tilde\pi}^\top (\tilde P - P), \quad Q = P -  1 \pi^\top \quad \text{and} \quad Z = (I-Q)^{-1} = \sum_{t=0}^\infty Q^t.
\]
Here $Z$ is know as the fundamental matrix of $P$. The sum over $t$ is well defined since (after we take out the only eigenvector of size $1$) the spectral radius of $Q$ is less than $1$. 

First notice that 
\begin{align}
v^\top Z 
&
=
\sum_{t=0}^\infty v^\top Q^t
\notag
\\
&
=
\sum_{t=0}^\infty {\tilde\pi}^\top (\tilde P - P) Q^t
\notag
\\
&
=
\sum_{t=0}^\infty 
	 \tilde{ \pi}^\top 
    	\left[
        	I - P -  1 \pi^\top 
        \right] Q^t + (\tilde{ \pi}^\top \bm 1)( \pi^\top Q^t)
\notag
\\
&
= 
\tilde{ \pi}^\top -  \pi  \, .
\label{PerbMC:1}
\end{align}
\noindent In the final inequality above we note that the term in square brackets is $Z^{-1}$, and that $\tilde{ \pi}^\top \bm 1 = 1$  and $\pi^\top Q^t = 0$ for $t\geq 1$.

Second notice that $ v^\top \bm 1=0$ so \
\[
 v^\top Q =  v^\top P -  v^\top \bm 1 =  v^\top P\, .
\]
Continuing inductively (noting that $\bm v^\top Q \bm 1 = 0$ also) gives that 
\begin{equation}\label{PerbMC:2}
  v^\top Q^t = v^\top P^t \, .
\end{equation}

Third, notice that
\begin{align}
||v||
= 
\frac{1}{2}
\sum_y \Big|
    \sum_x \tilde \pi_x (\tilde P_{xy}- P_{xy} )
\Big|
&
\leq 
\frac{1}{2}
\sum_y 
    \sum_x
     \pi_x
    |\tilde P_{xy}- P_{xy}|
\notag
\\
&
=
\frac{1}{2}
\sum_x
    \pi_x
    \sum_y
    |\tilde P_{xy}- P_{xy}|
\notag
\\
&
\leq 
\frac{1}{2}
\max_x 
    \sum_y
    |\tilde P_{xy}- P_{xy}|
\notag
\\
&
\leq 
\max_x 
|| \tilde P_{x\cdot } - P_{x,\cdot}  || \, .
\label{PerbMC:3}
\end{align}

Now putting everything together,
\begin{align}
    \left\| 
    	\tilde \pi^\top - \pi  
    \right\|
    &
    =
    \Big\|
        \sum_{t=0}^\infty {v}^\top Q^t
    \Big\|
    \tag{by \eqref{PerbMC:1}}
\\
&
\leq 
    \sum_{t=0}^\infty 
        \left\|
            {v}^\top Q^t
        \right\|
\notag
\\
&=
    \sum_{t=0}^\infty 
        \left\|
            {v}^\top P^t
        \right\|
\tag{by \eqref{PerbMC:2}}
\\
&
\leq 
    \sum_{t=0}^\infty 
        \left\|
            {v}^\top
        \right\|
        \rho(P^t)
\tag{by Definition}
\\
&
\leq 
    \sum_{t=0}^\infty 
        \left\|
            {v}^\top
        \right\|
        \rho(P)^t
\tag{By part b)}
\\
&
\leq
\frac{1}{1-\rho(P)}
\max_x 
|| \tilde P_{x\cdot } - P_{x,\cdot}  ||
\tag{by \eqref{PerbMC:3}}
\end{align}
which gives the required bound.
\Endproof

\begin{lemma}\label{AuxLem}
If $||v|| =1$ and $v^\top \bm 1 = 0$ then
\[
v = \sum_{x\neq y } \frac{u_{xy}}{2} (\bm e_ x - \bm e_y)
\]
for $u_{x,y}\geq 0$ and $\sum_{x\neq y} u_{xy} = 1$.
\end{lemma}
\Beginproof
	The result is proven by induction if $|\mathcal X|=2$ then $v^\top \bm 1=0$ implies $v= |v|(\bm e_1 - \bm e_2)/2$. So take $u_{1,2}=|v|$. 
We suppose the result is true for $|\mathcal X|=n$. Then for $|\mathcal X|=n+1$. We can write the vector $v = (v_0,v_1,...,v_n)$ as
\[
v^\top = 
\begin{pmatrix}0\\v_1+v_0\\ v_2\\\vdots\\v_n \end{pmatrix}
-
v_0\begin{pmatrix}1\\-1\\0\\{\vdots}\\0 \end{pmatrix} \, .
\]
We call the first vector above $\hat v$.
Wlog we can assumed $v_0<0$, $v_1 \geq -v_0$. (This can be achieved by permuting entries and, if necessary, multiplying the vector $v$ by $-1$.)
 Applying the induction hypothesis to $\hat v = \sum_{x\neq y} u_{x,y}(\bm e_x - \bm e_y)/2$ and setting $u_{0,1}=-2v_0$ gives an expression of the required form and also
\[
\sum_{x\neq y} u_{x,y} = u_{0,1} + \sum_{x,y\neq 0} u_{x,y}
= - 2v_0 + ||\hat v||_1
= - 2v_0 + (v_0 + v_1) + |v_2| + ... + |v_n| = ||v||_1 \, .
\]
In the above we note that because $v_1>-v_0>0$ it holds that $|v_1+v_0| = v_1 + v_0$
\Endproof

\begin{Repeatlemma}{RLem}
For a discounted program\\ a)
\[
\|
R(\cdot,P)
-
R(\cdot, \tilde P) \|_\infty 
\leq \frac{\beta r_{\max}}{(1-\beta)^2}
\| P- \tilde P\| .
\]
b)
\[
\| Q(\cdot,\cdot; P) - Q(\cdot. \cdot ; \tilde P)\|_\infty
\leq 
\frac{\beta r_{\max}}{(1-\beta)^2}
\| P- \tilde P\| .
\]
\end{Repeatlemma}
\Beginproof a) Note
\[
R(x,P) - R(x,\tilde P) 
=
\sum_{x\in\mathcal X} 
\left[
P^t - \tilde P^t
\right] r(x) 
\]
So by Theorem \ref{thrm:adiabatic}a) we have that 
\[
\| P^t - \tilde P^{t} \|
\leq t \| P - \tilde P \|
\]
Thus 
\[
\| R(\cdot ,P) - R(\cdot ,\tilde P)  \|
\leq 
\sum_{t=0}^T t \beta^t || P - \tilde P || r_{\max}
=
\frac{\beta}{(1-\beta)^2} \| P - \tilde P \| r_{\max}
\]
as required. This completes part a).

b) Now we define
\[
V(s ; P) :=\max_{a} Q(s,a;P)\,
\]
We note by definition 
\begin{equation}\label{Vinfty}
||V(\cdot ; P)||_\infty \leq \sum_{t=0}^\infty \beta^t r_{\max} \leq \frac{r_{\max}}{(1-\beta)}\, .	
\end{equation}
and
\begin{equation}\label{Vinf2}
\|
V(\cdot ; P) - V(\cdot ; \tilde P)
\|_\infty
\leq 
\|
Q(\cdot ; P) - Q(\cdot ; \tilde P)
\|_\infty \, .	
\end{equation}
(Note the above inequality was proven in \eqref{eqd} of Lemma \ref{lem:contraction}.)
Now for $x=(s,a)$
\begin{align}
 Q(x; P ) - Q(x;\tilde P)
&
=
\beta \left[
P V(x ; P)
-
\tilde P V(x ; \tilde P)
\right]
\notag
\\
& 
=
\beta \left[
P 
-
\tilde P 
\right]
V(x ;  P)
+
\beta \tilde P  \left[V(x ; P)
-
V(x ; \tilde P)
\right]\, .
\label{Qeqeq}
\end{align}
Thus applying \eqref{Vinfty} and \eqref{Vinf2} to \eqref{Qeqeq} gives
\begin{align*}
  || Q(\cdot;P)-Q(\cdot;\tilde P) ||_\infty 
\leq 
\frac{\beta r_{\max}}{1-\beta} || P - \tilde P|| 
+
\beta   || Q(\cdot;P)-Q(\cdot;\tilde P) ||_\infty 
\end{align*}
which rearranges to give the required bound.

\Endproof

\begin{Repeatlemma}{DiscRho}
	If $X_t$ is a time homogeneous Markov chain and we wish to evaluate 
\[
R(x) 
:= 
\mathbb E_x \left[
\sum_{t=0}^\infty \beta^t r(X_t)
\right]
\]
for $\beta \in (0,1)$ and $r:\mathcal X \rightarrow \mathbb R$ a bounded function, then for $\hat \beta \in ( \beta , 1)$ there exists a positive recurrent time-homogenous Markov chain $\hat x_t$ whose transition matrix, $P$, satisfies
\[
\rho(P) < \frac{\beta}{\hat \beta} <1
\qquad \text{
and}
\qquad
 \mathbb E_x \left[ \sum_{t=0}^\infty \hat \beta^t r(\hat x_t)\right]
=
\frac{1-\beta}{1-\hat \beta}
R(x) 
\,.
\]
\end{Repeatlemma}

\Beginproof
We let $\hat x_t$ be the Markov chain $X_t$ but restarted at $x$ with probability $\beta/\hat \beta$. Specifically, we let $G_k$, $k\in\mathbb Z_+$ be a sequence of random variables such that $G_0=0$ and, for $k \in \mathbb N$, $G_k - G_{k-1}$ is
 independent geometrically distributed random variables on $\mathbb Z_+$ with parameter $\beta / \hat \beta$.  
For each $k$, we let $X_t^k$ be an independent instance of the Markov chain $X_t$ started from $X_0=x$.
We define the Markov chain $\hat x_t$ by
\[
\hat x_t = 
	X^k_t\qquad   \text{for }  \quad G_{k} \leq t < G_{k+1},\quad k \in \mathbb Z_+ \,.
\]
Thus
\begin{align}
  \mathbb E_x \left[ 
 \sum_{t=0}^\infty
	\hat \beta^t r(\hat x_t) 
\right]
&
=
\mathbb E_x \left[
\sum_{k=0}^\infty 
\sum_{t=G_k}^{G_{k+1} -1} 
	\hat \beta^t r(\hat x_t) 
\right]
\notag
\\
&
=
\sum_{k=0}^\infty 
\mathbb E_x \left[
\hat \beta^{G_k}
\mathbb E
\Bigg[
	\sum_{t=G_k}^{G_{k+1}-1}
	\hat \beta^{t-G_k} r(\hat x_t) 
\bigg|
\hat x_0,..., \hat x_{G_k}
\Bigg]
\right]
\notag
\\
&
=\sum_{k=0}^\infty 
\mathbb E_x \left[
\hat \beta^{G_k}
\right]
\mathbb E_x \left[
\sum_{t=0}^{G_1}
\hat \beta^t r(X_t)
\right]\, .\label{MDP2Terms}
\end{align}
In the first two equalities above, we separate out terms and condition. In the third, inequality we apply the Markov property to the conditional expectation and then apply the definition that $\hat x_t = X_t$ for $t\leq G_1$. We now deal with the two terms in \eqref{MDP2Terms}.

 First
\begin{align}
  \mathbb E_x 
\left[
	\sum_{t=0}^{G_1-1}
		\hat \beta^t r( X_t)
\right]
=
\mathbb E_x \left[
	\sum_{t=0}^\infty	
		\hat \beta^t r(X_t) 
	\mathbb I [G_1 \geq t ]
\right]
=
\mathbb E_x 
\left[
	\sum_{t=0}^\infty
	\hat \beta ^t \left(
	\frac{\beta}{\hat \beta}
\right)^{t+1} 
r(X_t) 
\right]
=
\frac{\beta}{\hat \beta}
R(x)\,.\label{MDPfirst}
\end{align}

Second,
\begin{align}
	\mathbb E_x\left[
\hat \beta^{G_k}
\right]
= \mathbb E_x \left[
\hat \beta^{G_1}
\right]^k
=\left[\sum_{t=0}^\infty \hat \beta^t \left(1- \frac{\beta}{\hat \beta} \right) 
\left(\frac{\beta}{\hat \beta}\right)^t\right]^k
=
\left[\frac{1- \frac{\beta}{\hat \beta} }{1-\beta}\right]^k
\label{MDPsecond}
\end{align}

Thus applying \eqref{MDPfirst} and \eqref{MDPsecond} to \eqref{MDP2Terms} gives 
\begin{align*}
    \mathbb E_x \left[ 
 \sum_{t=0}^\infty
	\hat \beta^t r(\hat x_t) 
\right]
=
\frac{\beta}{\hat \beta}
R(x)
\sum_{k=0}^\infty
\left[\frac{1- \frac{\beta}{\hat \beta} }{1-\beta}\right]^k
=
\frac{\beta}{\hat \beta} R(x) \frac{1-\beta}{1- \beta - (1- \beta/\hat \beta)}
=
\frac{1-\beta}{1-\hat \beta} R(x)
\end{align*}
This gives the required expression for $R(x)$.

Now notice that the resulting transition matrix $P$ for $\hat x_t$ will transition to $x$ from any state with probability at least $\beta / \hat \beta$. Thus by Proposition \ref{prop:ergprop}c), we have that 
\[
1- \rho(P)
=
\min_{x_1,x_2} \sum_y \min \{ P_{x_1,y} , P_{x_2,y} \}
\geq \frac{\beta}{\hat \beta} 
\]
as required.
\Endproof

\section{Additional Proofs for Section \ref{sec:async}}\label{app:4}

The following lemmas are commonplace stochastic approximation results.

\begin{Repeatlemma}{lem:zcomp}
Suppose that $z_n$ is a positive real valued sequence such that 
\[
z_{n+1} \leq z_n (1 - a_n) + c_n
\]	
then 
\[
z_{n+1} \leq z_0 \prod_{k=0}^n (1-a_k)  +  \sum_{j=0}^n c_j \prod_{k=j+1}^n (1 - a_k)\, .
\]
\end{Repeatlemma}
\Beginproof
	The proof follows by repeated substitution 
\begin{align*}
z_{n+1} \leq z_n (1 - a_n) + c_n & \leq  ( z_{n-1} (1- a_{n-1} ) + c_{n-1} )  (1 - a_n) + c_n \\
& \leq  z_{n-1} (1- a_{n-1} ) (1 - a_n) +   c_n  + c_{n-1}  (1 - a_n) \\
&
\vdots
\\&\leq z_0 \prod_{k=0}^n (1-a_k)  +  \sum_{j=0}^n c_j \prod_{k=j+1}^n (1 - a_k)
\end{align*}

\Endproof

\begin{Repeatlemma}{lemma:alpha_sum}
	If we let $\alpha_t = t^{-\gamma}$, for $\gamma \in (0,\infty)$, we have  
	\[
	\frac{t^{1-\gamma} - s^{1-\gamma}}{ 1-\gamma }  
	\leq \sum_{n=s}^t  \frac{1}{n^{\gamma}} 
	\leq
	\frac{1}{s^{\gamma}} + \frac{t^{1-\gamma} - s^{1-\gamma} }{ 1-\gamma }  \, .
	\] 
where for $\gamma = 1$, we define
$\;
{(t^{1-\gamma} - 1)}/{(1-\gamma )}  := \log t
$.
\end{Repeatlemma}
\Beginproof
	This can be obtained by simple calculations.
	Since $n^{-\gamma}$ is decreasing for $\gamma > 0$, then
	\begin{align*}
	\sum_{n=s}^t  {n^{-\gamma}} \leq \frac{1}{s^\gamma} + \int_{s}^{t}{u^{-\gamma}} \mathrm{d} u = \frac{1}{s^\gamma} + \frac{t^{1-\gamma} - s^{1-\gamma}}{ 1-\gamma } \,,
	\end{align*}
	and
	\begin{align*}
	\sum_{n=s}^t  {n^{-\gamma}} \geq  \int_{s}^{t}{u^{-\gamma}} \mathrm{d} u =  \frac{t^{1-\gamma} - s^{1-\gamma}}{ 1-\gamma } \,.
	\end{align*}
\Endproof

\begin{Repeatlemma}{prodbound}
	For positive sequences $a_t$ and $b_t$ with $a_t \in(0,1)$ and $b_t$ decreasing
\[
\sum_{t=1}^T a_t b_t \prod_{s=t+1}^T (1-a_s) 
\leq 
b_{T/2} 
+
e^{-\sum_{t=T/2}^T a_t}
\sum_{t=1}^{T/2} a_t b_t
\]
Moreover if $a_t = \frac{C_a}{t^{\gamma_a}}$ and $b_t = \frac{C_b}{t^{\gamma_b}}$ for $\gamma_a \in(0,1)$ and $\gamma_b \geq 0$ then 
\[
\sum_{t=1}^T a_t b_t \prod_{s=t+1}^T (1-a_s) 
\leq 
\frac{D_{a,b}}{T^{\gamma_b}}
\, ,
\]
where $D_{a,b}$ is a constant depending on $C_a,C_b,\gamma_a$ and $\gamma_b$.
\end{Repeatlemma}

\Beginproof
\begin{align*}
\sum_{t=1}^T a_t b_t \prod_{s=t+1}^T (1-a_s) 
&
=
\sum_{t=T/2+1}^T a_t b_t \prod_{s=t+1}^T (1-a_s) 
+
\sum_{t=1}^{T/2} a_t b_t \prod_{s=t+1}^T (1-a_s) 
\\
&
\leq 
b_{T/2} \sum_{t=T/2+1}^T a_t \prod_{s=t+1}^T (1-a_s) 
+
e^{-\sum_{t=T/2}^T a_t}\sum_{t=1}^{T/2} a_t b_t 
\\
&
=
b_{T/2} \sum_{t=T/2+1}^T a_t \prod_{s=t+1}^T (1-a_s) 
+
e^{-\sum_{t=T/2}^T a_t}\sum_{t=1}^{T/2} a_t b_t
\\
&
= 
b_{T/2} \sum_{t=T/2+1}^T
\left[   \prod_{s=t+1}^T (1-a_s) 
-
 \prod_{s=t}^T (1-a_s) 
\right]
+
e^{-\sum_{t=T/2}^T a_t}\sum_{t=1}^{T/2} a_t b_t
\\
&
=
b_{T/2} \left[ 1-
 \prod_{s=T/2+1}^T (1-a_s) 
\right]
+
e^{-\sum_{t=T/2}^T a_t}\sum_{t=1}^{T/2} a_t b_t
\\
&
\leq
b_{T/2}
+
e^{-\sum_{t=T/2}^T a_t}\sum_{t=1}^{T/2} a_t b_t \, .
\end{align*}
If $a_t = \frac{C_a}{t^{\gamma_a}}$ and $b_t = \frac{C_b}{t^{\gamma_b}}$, 
\begin{align*}
 & b_{T/2}
+
e^{-\sum_{t=T/2}^T a_t}\sum_{t=1}^{T/2} a_t b_t
\\
&
=
\frac{2^{\gamma_b}C_b}{T^{\gamma_b}}
+
\exp\left\{
- \sum_{T/2}^T \frac{C_a}{ {t}^{\gamma_a}}
\right\}
\sum_{t=1}^{T/2} 
	\frac{C_a C_b  }{t^{\gamma_a + \gamma_b}}
\\
&
=
\frac{2^{\gamma_b}C_b}{T^{\gamma_b}}
+
\exp\left\{
-C_a \left[ T^{1-\gamma_a} - \frac{T^{1-\gamma_a}}{2^{1-\gamma_a}} \right]
\right\}
\left[ 
C_a C_b \frac{ T^{1-\gamma_a - \gamma_b} - 1}{1- \gamma_a - \gamma_b}
+
C_a C_b 
\right]
\\
&
=
\frac{2^{\gamma_b}C_b}{T^{\gamma_b}}
+
\frac{1}{T^{\gamma_b}}
\left[
\exp\left\{
-C_a \left(1 - \frac{1}{2^{1-\gamma_a}} \right) T^{1-\gamma_a}
\right\}
\left(
C_a C_b
 \frac{ T^{1-\gamma_a} - T^{\gamma_b}}{1- \gamma_a - \gamma_b}
+
C_a C_b T^{\gamma_b}
\right)
\right]\, .
\end{align*}
It should be clear that the term in the square brackets goes to zero thus giving the result.
\Endproof

\begin{Repeatlemma}{Lem:Atoa}
For any sequence $A_T$ we can write $A_T$ as
\begin{equation}\label{eq:at}
A_T 
=
\sum_{t=1}^T
a_t \alpha_t
\prod_{s=t+1}^T (1-\alpha_s)	
\end{equation}
where 
\[
a_t = \frac{A_t - A_{t-1}}{\alpha_t} + A_{t-1}\,.
\]
Thus if $A_t = C_A/ t^{\gamma_A}$ and $\alpha_t = C_\alpha/ t^{\gamma_\alpha}$ for $\gamma_{\alpha} \in (0,1]$ then 
\[
a_t \leq \frac{D_{\alpha, A} }{t^{\gamma_A}}
\]
for $D_{\alpha,A}$ a positive constant depending on $\gamma_{\alpha}$, $\gamma_A$, $C_{\alpha}$, $C_A$.
\end{Repeatlemma}

\Beginproof
Notice that $A_t$ obeys the recursion
\[
A_t = \alpha_t a_t + (1-\alpha_t) A_{t-1}\,. 
\]
Now apply Lemma \ref{lem:zcomp} to obtain \eqref{eq:at}.

If $A_t = C_A/ t^{\gamma_A}$ then by convexity for $t\geq 2$
\[
A_t - A_{t-1} = \frac{C_A}{ t^{\gamma_A} } - \frac{C_A}{ (t-1)^{\gamma_A}}  \leq \frac{C_A \gamma_A}{(t-1)^{\gamma_A+1}}
=
\frac{C_A \gamma_A}{t^{\gamma_A+1}(1-1/t)^{\gamma_A+1}}
\leq 
\frac{C_A \gamma_A 2^{\gamma_A+1}}{t^{\gamma_A+1}} \, .
\]
Applying this bound to $a_t$ gives
\[
a_t 
=
\frac{A_t- A_{t-1}}{\alpha_t} + A_{t-1} 
\leq 
\frac{C_A \gamma_A 2^{\gamma_A+1}/C_{\alpha}}{t^{\gamma_A+1-\gamma_{\alpha}}}
+ 
\frac{C_\alpha 2^\gamma_\alpha}{ t^{\gamma_\alpha}}
\leq
\frac{C_A \gamma_A 2^{\gamma_A+1}/C_{\alpha}+C_\alpha 2^\gamma_\alpha}{t^{\gamma_A}}
\, ,
\]
where the final inequality above holds since $1-\gamma_{\alpha} \geq 0$
\Endproof

\begin{Repeatlemma}{zboundLem}For positive sequences $z_t$, $\alpha_t$ if
\[
z_{t+1}
\leq 
z_0 \prod_{s=1}^t (1-\alpha_s) 
+
\sum_{s=1}^t \beta  z_s 
\prod_{u=s+1}^t (1-\alpha_u)
+
\sum_{s=1}^t c_s 
\prod_{u=s+1}^t (1-\alpha_u)
\]
then $z_t \leq \tilde z_t$ where $\tilde z_t$ solves the recursion
\[
\tilde z_{t+1} = (1-\alpha_t (1-\beta)) \tilde z_t + c_t
\]
with $\tilde z_0\geq  z_0$.
\end{Repeatlemma}

\Beginproof
	The result holds by induction. 
By assumption, $z_0 \leq \tilde z_0$.
Suppose the induction hypothesis: that $z_s \leq \tilde z_s$ for all $s\leq t$.
Expanding the recursion for $\tilde z_t$. 
\[
\tilde z_{t+1} = \tilde z_0 \prod_{s=1}^t (1-\alpha_s) 
+
\sum_{s=1}^T \beta  \tilde z_s 
\prod_{u=s+1}^T (1-\alpha_u)
+
\sum_{s=1}^t c_s 
\prod_{u=s+1}^t (1-\alpha_u)
\]
Thus by assumption, the inductive hypothesis and the above bound:
\begin{align*}
	z_{t+1}
&\leq 
z_0 \prod_{s=1}^t (1-\alpha_s) 
+
\sum_{s=1}^t \beta  z_s 
\prod_{u=s+1}^t (1-\alpha_u)
+
\sum_{s=1}^t c_s 
\prod_{u=s+1}^t (1-\alpha_u)
\\
&
\leq
\tilde z_0 \prod_{s=1}^t (1-\alpha_s) 
+
\sum_{s=1}^T \beta  \tilde z_s 
\prod_{u=s+1}^T (1-\alpha_u)
+
\sum_{s=1}^t c_s 
\prod_{u=s+1}^t (1-\alpha_u)
\\
&= \tilde z_{t+1} \, .
\end{align*}
as required.
\Endproof

%
%

The following Lemma is a slight variant of Lemma 13 from \cite{qu2020finite}.

\begin{Repeatlemma}{AHbound}
	If $\epsilon_t$ is adapted with $\mathbb E[ \epsilon_t | \mathcal F_{t-\tau}]=0$ and $|\epsilon_t| \leq \epsilon_{\max}$ then with probability greater than $1-\delta$  it holds that, for all $t$ with $\tau \leq t \leq T$,
\[
\left| 
\sum_{s=\tau}^t 
	\alpha_s 
\epsilon_s 
\prod_{u=s+1}^t (1-\alpha_s \pi_s)
\right| 
\leq 
\sqrt{
2  \tau
\left[\sum_{s=1}^t \epsilon^2_{\max} \alpha_s^2
\prod_{u=s+1}^t 
(1-\alpha_s\pi_s)^2
\right]
 \log \bigg(\frac{2\tau T}{\delta}\bigg)
} 
\]
\end{Repeatlemma}
\Beginproof
We let $z_t$ be equal to the upper-bound in that we wish to prove. Specifically, we define 
\[
z_t := \sqrt{
2  \tau
\left[\sum_{s=1}^t \epsilon^2_{\max} \alpha_s^2
\prod_{u=s+1}^t 
(1-\alpha_s\pi_s)^2
\right]
 \log \bigg(\frac{2\tau T}{\delta}\bigg)
}  \, .
\]
We let $E_t$ be the term we wish to bound: 
\[
E_t := \sum_{s=\tau}^t 
	\alpha_s 
\epsilon_s 
\prod_{u=s+1}^t (1-\alpha_s \pi_s)\,.
\]
Notice that 
\[
E_t = \alpha_t \epsilon_t + (1-\alpha_t\pi_s) E_{t-1} \, .
\]
If we define 
\[
z'_t= 
\frac{z_t}{\prod_{s=\tau}^t (1-\alpha_s \pi_s)}\,,
\qquad
M_t := \frac{E_t}{\prod_{s=\tau}^t (1-\alpha_s \pi_s)}
\qquad 
\text{and}
\qquad 
\eta_t :=  \frac{\alpha_t}{\prod_{s=\tau}^t (1-\alpha_s \pi_s)}
\]
then 
\[
M_t = \eta_t \epsilon_t + M_{t-1} 
=
\sum_{s= \tau}^t \eta_s \epsilon_s \, .
\]
Notice if  we define 
\[
M_t^{(l)}
=
\sum_{k \in \mathbb N: k \tau+ l \leq t}
\alpha_{k\tau+ l}
\epsilon_{k\tau+l}
\]
then $M^{(l)}_{k\tau+l}$, $k\in\mathbb Z_+$ is a martingale with respect to the filtration $\{ \mathcal F_{k\tau+l} \}_{k\in \mathbb Z_+}$ and 
\[
M_t = \sum_{l=0}^{\tau-1} 
M^{(l)}_t\, .
\]
Also we define $z^{(l)}_t$ to be a deterministic sequence of positive numbers such that $\sum_{l=0}^{\tau-1} z^{(l)}_t \leq z'_t$.

Now applying the definitions above, a union bound and then the Azuma-Hoeffding Inequality \cite[p237]{williams1991probability}, for any deterministic sequence $z_t$ we have that 
\begin{align*}
	\mathbb P \Big( \exists t\leq T, | E_t| \geq z_t \Big)
&
=
	\mathbb P \Big( \exists t\leq T, | M_t| \geq z'_t \Big)
\\
& \leq 
\sum_{t=1}^T
\sum_{l=0}^{\tau-1}
\mathbb P( | M_t^{(l)} |\geq z^{(l)}_t) 
\\
&
\leq 
\sum_{t=1}^T
\sum_{l=0}^{\tau-1}
2 \exp\left\{
-\frac{1}{2}
\frac{(z^{(l)}_t)^2}{\sum_{k:k\tau+l \leq t} \epsilon_{\max}^2\eta^2_{k\tau+l}}
\right\}\, .
\end{align*}
To prove the result it is sufficient to prove that each term in sum above is less that $\delta/(T\tau)$.

Notice it holds that
\[
2 \exp\left\{-
\frac{1}{2}
\frac{(z^{(l)}_t)^2}{\sum_{k:k\tau+l \leq t} \epsilon_{\max}^2 \eta_{k\tau+l}^2}
\right\}
\leq \frac{\delta}{\tau T} 
\]
for $z^{(l)}_t$ such that
\[
z^{(l)}_t :=  \sqrt{2 \bigg[\sum_{k : k \tau + l \leq t } \epsilon_{\max}^2\eta^2_{k\tau +l } \bigg]\log \left(\frac{2T \tau }{\delta}\right)}\, .
\]
Thus notice by Jensen's Inequality it holds that
\begin{align*}
\sum_{l=0}^{\tau-1} z^{(l)}_t 
\leq 
&
\sqrt{
2 \tau
\bigg[
\sum_{l=0}^{\tau-1}
\sum_{k : k \tau + l \leq t}
\epsilon^2_{\max} \eta^2_{k\tau +l } \bigg]\log \left(\frac{2T \tau }{\delta}\right)}
\\
=
&
\sqrt{
2 \tau
\bigg[
\sum_{s=\tau}^t
\epsilon^2_{\max} \eta^2_{s} \bigg]\log \left(\frac{2T \tau }{\delta}\right)}
=: z'_t \, .
\end{align*}
Finally, we recover $z_t$ and the required bound by multiplying $z'_t$ as above by $\prod_{s=1}^t(1-\alpha_s \pi_s)$.
Specifically, we note that
\begin{align*}
  z_t
&
= 
z'_t \prod_{s=\tau}^t (1-\alpha_s \pi_s)^{-1}
\\
&
=
\sqrt{
2\tau 
\left[
\sum_{s=\tau}^t \epsilon^2_{\max}\eta_s
\prod_{u=1}^t(1-\alpha_s \pi_s)^{-2}
\right]
\log \Big(\frac{2T\tau}{\delta}\Big)
}
\\
&
\leq
\sqrt{
2\tau 
\left[
\sum_{s=1}^t \epsilon^2_{\max} \alpha_s
\prod_{u=s+1}^t(1-\alpha_s \pi_s)^{2}
\right]
\log \Big(\frac{2T\tau}{\delta}\Big)
}\,.
\end{align*}

%
%
%
\Endproof

\begin{Repeatlemma}{lem7} 
For $\tau = 8 \frac{\log T }{|\log \rho|}$ and $t$ such that $\tau \leq t \leq T$ it holds that 
\[
\|
\mathbb P( \hat x_t = \cdot | \mathcal F_{t-\tau})
-
\pi^{(t)}(\cdot)
\|
\leq 
\frac{1}{(1-\rho)^2}\frac{D_P}{t^{\gamma_P}} 
+
\frac{4}{(1-\rho)^2}
\frac{\log T}{T^4}
+
\frac{1}{T^4}\, ,
\]
where $D_P =C_P 2^{\gamma_P}$.
\end{Repeatlemma}
\Beginproof
By the Adiabatic Theorem, Theorem \ref{thrm:adiabatic}c), it holds that 
for $t\geq \tau$,
\begin{align*}
\| \mathbb P( \hat x_t = \cdot | \mathcal F_{t-\tau})
-
\pi^{(t)}(\cdot) \|
\leq
& 
\phi_{t-\tau/2} \frac{\rho}{(1-\rho)^2}
+
\frac{\rho^{\tau/2+1}}{1-\rho}
\sum_{s=1}^{\tau/2} \phi_{t-\tau+s}
+
\rho^{\tau}
\\
=
&
\phi_{t-\tau/2} \frac{\rho}{(1-\rho)^2}
+
\frac{\tau \rho^{\tau/2+1}	}{1-\rho}\phi_{t-\tau} 
+
\rho^\tau \, 
\end{align*}
where $\phi_t=C_P/t^{\gamma_{P}}$ is as given in \eqref{RP}. Substituting in this and the value of $\tau$ gives
\begin{align*}
	\| \mathbb P( \hat x_t = \cdot | \mathcal F_{t-\tau})
-
\pi^{(t)}(\cdot) \| 
&\leq
\frac{C_P}{(t-\tau/2)^{\gamma_P}}
\frac{\rho}{(1-\rho)^2}
+
\frac{8\log T}{T^4 (1-\rho) | \log \rho|}
+ \frac{1}{T^4}
\end{align*}
Finally, noting that $(t-\tau/2)^{\gamma_P} = t^{\gamma_P} (1- \tau/(2t))^{\gamma_P} \geq t^{\gamma_P} 2^{-\gamma_P}$ and $|\log \rho| \geq 1- \rho$ for $\rho<1$, gives the required bound:
\[
\| \mathbb P(\hat x_t = \cdot | \mathcal F_{t-\tau}) - \pi^{(t)}(\cdot)\|
\leq \frac{1}{(1-\rho)^2}\frac{D_P}{t^{\gamma_P}} 
+
\frac{4}{(1-\rho)^2}
\frac{\log T}{T^4}
+
\frac{1}{T^4}\,.
\]
where $D_P=C_P 2^{\gamma_P}$.
\Endproof

\begin{Repeatlemma}{Lem:Bound}
	The sequence $R_t$ defined in \eqref{SA:update} and $F_tR_t$
 are bounded in $t$.
\end{Repeatlemma}
\Beginproof
Note that
\begin{align*}
	||R_{t+1} ||_\infty
&
\leq 
(1-\alpha_t)||R_t ||_\infty + \alpha_t ||F_tR_t||_\infty + \alpha_t || \epsilon_t||_\infty 
\\
&
\leq
(1- \alpha_t )||R_t ||_\infty + \beta \alpha_t
||R_t ||_\infty 
+ \alpha_t F_{\max}
+ \alpha_t \epsilon_{\max}
\\
&
\leq
||R_t ||_\infty 
+ \alpha_t \left[ F_{\max} + \epsilon_{\max}
-(1-\beta) || R_t||_\infty\right]
\end{align*}
Thus, given the term in square brackets above, we see that if  
\[
||R_t||_\infty \leq \frac{F_{\max} + \epsilon_{\max} }{1-\beta} 
\]
then 
\[
||R_{t+1} || \leq \frac{F_{\max} + \epsilon_{\max} }{1-\beta} \,.
\]
Since $||R_0||=0$ the result bound above must hold for all $t\in\mathbb Z_+$. 

Applying the assumption that $||F_t R ||_{\infty} \leq ||R||_{\infty}+ C$, for all $R$, shows that $F_tR_t$ is bounded also.
\Endproof

\end{document}